\documentclass[journal]{IEEEtran}

\usepackage{comment}
\usepackage{epsfig}
\usepackage{graphicx}
\usepackage{amsmath}
\usepackage{amssymb}
\usepackage{comment}
\usepackage{multirow}
\usepackage{bbding}

\usepackage{hyperref}
\hypersetup{hidelinks}

\usepackage{booktabs}
\usepackage{array, caption, threeparttable}
\usepackage{caption}
\usepackage{subfigure}
\usepackage{algorithm}
\usepackage{listings}
\usepackage{color}
\usepackage{mathtools}

\usepackage{todonotes}
\usepackage{gensymb}

\begin{document}

\title{Advancing Generalizable Remote Physiological Measurement through the Integration of Explicit and Implicit Prior Knowledge}

\author{Yuting Zhang, Hao Lu, $\text{Xin Liu}^{\dagger}$,~\IEEEmembership{Senior Member,~IEEE}, Yingcong Chen, $\text{Kaishun Wu}^{\dagger}$,~\IEEEmembership{Fellow,~IEEE} \\

\thanks{Manuscript received March 10, 2024; $\dagger$ Corresponding author: Xin Liu (email: linuxsino@gmail.com) and Kaishun Wu (email: wuks@hkust-gz.edu.cn}

\thanks{Yuting Zhang, Hao Lu, Yingcong Chen and Kaishun Wu are with the Information Hub, the Hong Kong University of Science and Technology (Guangzhou), Guangzhou 511400, China. }

\thanks{Xin Liu is with Computer Vision and Pattern Recognition Laboratory, School of Engineering Science, Lappeenranta-Lahti University of Technology LUT, Lappeenranta 53850, Finland.} 
}

\markboth{IEEE TRANSACTIONS ON IMAGE PROCESSING}%
{Shell \MakeLowercase{\textit{et al.}}: Bare Demo of IEEEtran.cls for IEEE Journals}

\maketitle

\begin{abstract}
Remote photoplethysmography (rPPG) is a promising technology that captures physiological signals from face videos, with potential applications in medical health, emotional computing, and biosecurity recognition. The demand for rPPG tasks has expanded from demonstrating good performance on intra-dataset testing to cross-dataset testing (i.e., domain generalization). However, most existing methods have overlooked the prior knowledge of rPPG, resulting in poor generalization ability. In this paper, we propose a novel framework that simultaneously utilizes explicit and implicit prior knowledge in the rPPG task. Specifically, we systematically analyze the causes of noise sources (e.g., different camera, lighting, skin types, and movement) across different domains and incorporate these prior knowledge into the network. Additionally, we leverage a two-branch network to disentangle the physiological feature distribution from noises through implicit label correlation. Our extensive experiments demonstrate that the proposed method not only outperforms state-of-the-art methods on RGB cross-dataset evaluation but also generalizes well from RGB datasets to NIR datasets. The code is available at \href{https://github.com/keke-nice/Greip}{https://github.com/keke-nice/Greip}.
\end{abstract}


\IEEEpeerreviewmaketitle

\section{Introduction}

\begin{figure}[!t]
\begin{center}
\includegraphics[scale=0.9]{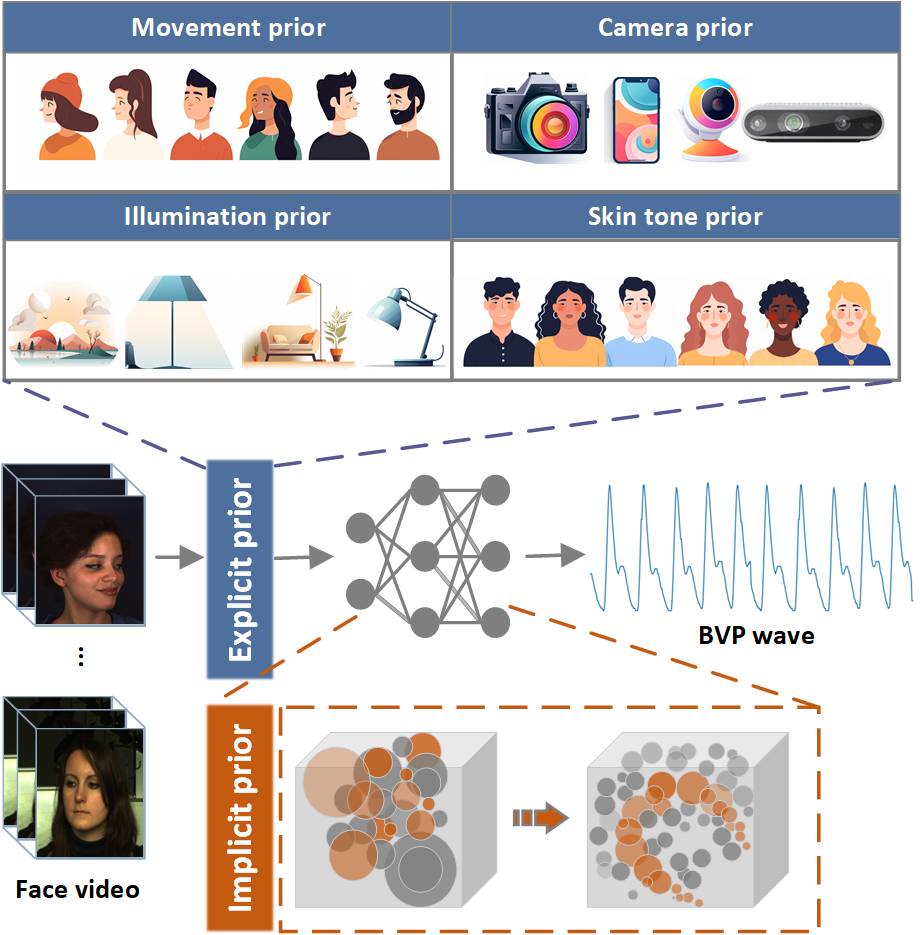}
\caption{The framework of Greip to utilize the explicit and implicit prior knowledge. Firstly, we incorporate explicit priors into the network in a unified augmentation way. Subsequently, we utilize the continuous implicit prior of rPPG labels to impose constraints on the rPPG features and noise within the network, which effectively transforms the network from a chaotic feature space into a distinguishable and continuous one.}
\label{fig:Intro}
\end{center}
\vspace{-10mm}
\end{figure}

\begin{table*}[!t] 
\small
\centering
\caption{The application of prior knowledge to different methods.}
\begin{tabular}{lccccc} 
\toprule  
\textbf{Method} &  \textbf{Movement prior}& \textbf{Camera prior} & \textbf{Illumination prior}& \textbf{Skin tone prior} & \textbf{rPPG feature prior}\\
\midrule
\textbf{GREEN~\cite{GREEN}} & \CheckmarkBold & \XSolidBrush &\XSolidBrush&\XSolidBrush & \XSolidBrush\\
\textbf{Poh2010~\cite{poh2010non}} &\CheckmarkBold& \XSolidBrush& \XSolidBrush&\XSolidBrush & \XSolidBrush\\
\textbf{Wang2017\cite{wang2017amplitude}} &\XSolidBrush& \XSolidBrush &\CheckmarkBold &\XSolidBrush & \XSolidBrush\\

\textbf{CHROM~\cite{CHROM}} & \CheckmarkBold & \XSolidBrush &\XSolidBrush & \XSolidBrush & \XSolidBrush\\

\textbf{SLF-RPM~\cite{wang2022self}} & \XSolidBrush & \XSolidBrush& \XSolidBrush & \XSolidBrush & \CheckmarkBold\\

\textbf{SIMPER~\cite{yang2022simper}} &\XSolidBrush & \XSolidBrush & \XSolidBrush & \XSolidBrush & \CheckmarkBold\\

\textbf{rPPG-MAE~\cite{liu2023rppgMAE}} &  \CheckmarkBold & \XSolidBrush & \XSolidBrush &\XSolidBrush &\CheckmarkBold\\

\textbf{Contrast-phys+~\cite{sun2024contrast}} & \XSolidBrush& \XSolidBrush & \XSolidBrush & \XSolidBrush & \CheckmarkBold\\

\textbf{Kurihara21~\cite{kurihara2021non}} & \XSolidBrush & \CheckmarkBold& \XSolidBrush & \XSolidBrush & \XSolidBrush\\
\midrule
\textbf{Greip (Ours)} & \CheckmarkBold & \CheckmarkBold & \CheckmarkBold & \CheckmarkBold  & \CheckmarkBold\\
\bottomrule 
\end{tabular} 
\label{tab:prior}
\vspace{-3mm}
\end{table*} 


\IEEEPARstart{I}{n} 2008, Verkruysse and his colleagues were the pioneers in proposing the use of remote photoplethysmography (rPPG) technology to measure physiological indicators~\cite{GREEN}, marking a transition in the field of physiological monitoring from traditional contact methods to non-contact methods. rPPG technology can extract blood volume pulse (BVP) from face videos, analyzing the periodic changes in skin light absorption caused by heartbeats. This technology can detect vital signs such as heart rate (HR), heart rate variability (HRV), and respiration frequency (RF), which are important indicators of the human body's sympathetic activation level. Additionally, rPPG technology obtained from face measurement can be used for tasks such as emotion computing~\cite{yang2021non_emotion_computing, huang2021spatio_emotion_computing, mcduff2014remote_emotion_computing}, video fraud detection~\cite{speth2021deception}, and biometric security~\cite{yu2021transrppg, qi2020deeprhythm}.

The methodology for rPPG tasks has evolved significantly over the years, marking a shift from conventional hand-crafted techniques~\cite{ Traditional_balakrishnan2013detecting,Traditional_lam2015robust, Traditional_li2014remote, Traditional_poh2010advancements,2SR, CHROM, POS, PVB} to deep learning-based approaches~\cite{bvpnet, physnet, deepphys, yu2022physformer, sun2022contrast, dual-GAN, TS-CAN,liu2023rppgMAE,liu2023robust}. Throughout this transformative journey, a plethora of novel techniques and strategies have emerged, be it in the enhancement of backbone networks~\cite{yu2022physformer, yu2023physformer++, liu2023efficientphys} or the refinement of training methodologies~\cite{dual-GAN}. This wave of innovation has empowered individual datasets to reach impressive levels of precision~\cite{bvpnet, physnet, deepphys, yu2022physformer, sun2022contrast,dual-GAN}. However, such high-precision achievements on isolated datasets do not align with the escalating requirements of rPPG applications. The crux of the issue lies in the pronounced discrepancies in predictive accuracy among various rPPG datasets, discrepancies that persist even under well-controlled experimental conditions, and are exacerbated in the unpredictable and multifaceted environments of real-world applications. This gap thus presents a significant hurdle for the practical deployment and broader dissemination of rPPG technology.

The pursuit of robust cross-dataset performance in rPPG analysis has recently garnered considerable interest within the research community~\cite{sun2022contrast, liu2023rppgMAE,chung2022domain1, NEST,sun2023resolve,du2023dual}. A number of studies~\cite{chung2022domain1,du2023dual,liu2023robust} have sought to extricate the intrinsic physiological signals from confounding domain-specific noise, aiming to enhance the universality and reliability of rPPG measurements. Specifically, the NEST-rPPG framework~\cite{NEST} introduces an innovative approach to training, designed to maximize feature space coverage, thereby bolstering the model's ability to generalize across different domains. This method not only demonstrates improved performance in unseen test environments but also establishes a comprehensive domain generalization protocol tailored for the rPPG task, paving the way for future advancements in the field.


While previous research has made strides in rPPG analysis \cite{kurihara2021non, NEST,chung2022domain1}, there remains a lack of systematic examination into the diverse noise sources originating from different domains, which is arguably a critical factor for enhancing model generalization. Table \ref{tab:prior} offers a comparative analysis between the prior knowledge utilized in existing methodologies and that which is incorporated within the framework proposed in this paper. Noise can emanate from a multitude of variables, such as the subject's physical movements, skin type, the camera's specifications, and the variability in lighting conditions. These variables introduce complexities that can be characterized and modeled as explicit prior knowledge, a concept graphically depicted in Figure~\ref{fig:Intro} (Explicit Prior). This visualization prompts the pivotal inquiry of how to effectively integrate such explicit prior knowledge into the architecture of deep learning models to fortify their generalization capabilities across disparate conditions.

Furthermore, the rPPG task is inherently a regression challenge, wherein the features should exhibit a continuum of change that mirrors the progressive nature of their corresponding labels. This concept, termed implicit prior knowledge, is illustrated in Figure~\ref{fig:Intro} (Implicit Prior). The recognition and utilization of this implicit continuum can serve as a powerful lever in calibrating the model to not only recognize but also adapt to the nuanced variations inherent in physiological data. By harnessing both explicit and implicit prior knowledge, we can significantly advance the generalization performance of deep learning models in rPPG tasks, consequently improving their robustness and efficacy in real-world applications.

In this paper, we aim to improve the \textbf{G}eneralization performance of \textbf{r}emote physiological measurement through \textbf{e}xplicit and \textbf{i}mplicit \textbf{p}riors (\textbf{Greip}). Specifically, we systematically summarize the explicit priors in the rPPG dataset and classified them into four categories: camera, motion, illumination, and skin color. Concurrently, we propose corresponding strategies to inject these explicit prior knowledge into the neural network. Regarding the implicit prior, we employ a dual-stream network to learn the rPPG feature distribution and noise distribution. We further constrain the rPPG feature distribution using the heart rate's continuity distribution. Additionally, we introduce an orthogonal constraint between the noise space and the rPPG space to maximize the implicit noise content. The acquired implicit noise distribution, when combined with rPPG features, can still provide reliable prediction performance. By combining explicit and implicit priors, the proposed model can achieve better generalization performance to deal with unknown data and situations. Notably, our model can even get the network to learn heart rate across modes by infusing all kinds of prior knowledge. To summarize, the contributions are listed as follows:
\begin{itemize}
    \item We propose Greip framework to improve the generalization performance of model, which can utilize both explicit and implicit prior knowledge. 
    
    \item In terms of explicit priors, we systematically summarized and classified the explicit priors of camera, motion, illumination and skin color existing in the rPPG dataset, and proposed corresponding coping strategies. 
    
    \item In terms of implicit prior, we disentangle more genuine rPPG feature distribution from various noises, which is based on label relationship.

    \item To the best of our knowledge, this study represents the pioneering instance of achieving cross-mode generalization (from RGB video to near-infrared video).

    \item The extended experiments conducted on self-supervised and semi-supervised learning rigorously validate the Greip method's exceptional generalization capabilities.
    
\end{itemize}

\vspace{-3mm}
\section{Related Work}
\label{sec:relatedwork}

\subsection{Remote Physiological Measurement}
rPPG is a non-invasive method for collecting physiological data by analyzing skin color changes in facial videos. Its evolution has progressed from traditional methods to supervised learning techniques, and now to self-supervised learning approaches. Traditional rPPG methods used techniques like blind source separation (BSS)~\cite{Traditional_lam2015robust, Traditional_li2014remote, Traditional_poh2010advancements} or the creation of projection planes/subplanes~\cite{CHROM, POS, 2SR, PVB}. These methods created special color spaces to extract rPPG signals and separate noise. While effective at improving the pulse frequency's signal-to-noise ratio in simple scenarios, they have limitations in more complex, less controlled scenes.
The advent of deep learning has led to a surge of supervised learning techniques in the rPPG field. This has resulted in a progressive evolution of the backbone network, transitioning from Convolutional Neural Networks (CNNs)~\cite{deepphys, TS-CAN, rhythmnet, HR-CNN, physnet}, to Generative Adversarial Networks (GANs)~\cite{song2021pulsegan, dual-GAN}, and now to Transformers~\cite{liu2023efficientphys, yu2022physformer, yu2023physformer++}. 
However, supervised learning techniques pose a significant challenge due to their requirement for extensive labeled datasets. The emergence of self-supervised learning methodologies~\cite{gideon2021way, yang2022simper, wang2022self, sun2022contrast, liu2023rppgMAE, speth2023non} has offered a solution, easing the difficulties associated with label acquisition in the rPPG field. 
Yet, these methods often overlook the practical challenges associated with obtaining labels and even test data in real-world applications. In the context of the network, the test sample essentially represents an unfamiliar domain. Historically, many existing techniques have concentrated on improving performance within specific datasets, inadvertently limiting their capacity to generalize across multiple datasets. Going forward, our objective is to boost domain generalization performance and address the challenges linked to the practical implementation of rPPG technology.

\vspace{-3mm}
\subsection{Domain generalization}
Domain generalization (DG) aims to train a model on one or multiple source domains to generalize to an unseen domain. The primary solutions fall into three categories: data manipulation~\cite{data_man_shankar2018generalizing, data_man_yue2019domain}, representation learning~\cite{AD, groupDRO,wang2024inter}, and meta-learning~\cite{meta_lv2022causality, sankaranarayanan2023meta}. The aforementioned methods primarily target general tasks like image classification and segmentation and are not necessarily customized for rPPG tasks. This is mainly due to the absence of distinct stylistic characteristics between different domains of rPPG tasks. Recognizing this gap, a number of domain generalization methods have been developed specifically for rPPG tasks. 
Initially, ~\cite{chung2022domain1} endeavored to segregate domain-invariant features across different domains, a common approach in domain generalization methods for other tasks. Specifically, it sought to decouple rPPG, identity, and domain characteristics. However, for rPPG tasks, it's challenging to directly abstract domain change characteristics into identity and domain characteristics.
Recognizing this issue, ~\cite{NEST} took a different approach, starting with the feature space. It proposed maximizing the coverage of the feature space during training, thus decreasing the likelihood of unoptimized feature activation during inference. 
Up until this point, existing methods hadn't directly addressed the problem of domain generalization specific to the characteristics of the rPPG domain. Indeed, the generalization performance of rPPG tasks can be influenced by a variety of factors, including camera type, lighting conditions, skin color, and motion.
In response to this, we conducted a systematic analysis of these noise sources and devised corresponding data augmentation strategies for different explicit noise priors. Simultaneously, we constructed an implicit noise distribution to account for unknown noise sources.

\section{Methodology}
\label{sec:method}

Given the input face video clip $x\in \mathcal{X}$ and the ground-truth $y\in \mathcal{Y}$ (i.e., HR, BVP signal), the general goal is to learn $f$: $\mathcal{X} \rightarrow \mathcal{Y}$, which can also be formulate as $\mathcal{P}(\mathcal{Y}|\mathcal{X}) \propto \mathcal{P}(\mathcal{X}|\mathcal{Y})\cdot\mathcal{P}(\mathcal{Y})$ in a Bayes theorem way. For domain generalization problem, we should migrate domain-specific noises $z_{n}$ (e.g., different illumination, camera parameters, motions, etc) and preserve domain-agnostic features $z_{phy}$ (i.e., physiological information). Thus, the $\mathcal{P}(\mathcal{Y}|\mathcal{X})$ can be further converted into the following formula:
\begin{equation}
    \begin{aligned}
        \mathcal{P}(y|x) &= \frac{\mathcal{P}(x|y)}{\mathcal{P}(x)}\cdot{\color{gray}\mathcal{P}(y)}\\
        &= \frac{\mathcal{P}(z_{phy}, z_n|y)}{\mathcal{P}(z_{phy}, z_n)}\cdot{\color{gray}\mathcal{P}(y)}\\
        &= {\color{gray}\underbrace{\frac{\mathcal{P}(z_{phy}|y)}{\mathcal{P}(z_{phy})}}\limits_{roubust}}\cdot\underbrace{\frac{\mathcal{P}(z_n|y, z_{phy})}{\mathcal{P}(z_n|z_{phy})}}\limits_{prejudiced}\cdot{\color{gray}\underbrace{\mathcal{P}(y)}\limits_{gt}},
    \end{aligned}
\end{equation}
where $\frac{\mathcal{P}(z_{phy}|y)}{\mathcal{P}(z_{phy})}$ is a robust relationship between the ground-truth and physiological information; $\frac{\mathcal{P}(z_n|y, z_{phy})}{\mathcal{P}(z_n|z_{phy})}$ is a bias term introduced by overfitting various noises in the source domain; $\mathcal{P}(y)$ reflect the ground-truth distribution in the source domain. In this paper, we focus on mitigating the negative effects of the domain-specific bias $\frac{\mathcal{P}(z_n|y, z_{phy})}{\mathcal{P}(z_n|z_{phy})}$.

Moreover, assuming the mutual independence and conditional independence across different noises ($n_i$)~\cite{tang2022invariant}, the $\frac{\mathcal{P}(z_n|y, z_{phy})}{\mathcal{P}(z_n|z_{phy})}$ can be further written as:
\begin{equation}
    \frac{\mathcal{P}(z_n|y, z_{phy})}{\mathcal{P}(z_n|z_{phy})} = \prod_{n_i\in n} \frac{\mathcal{P}(z_{n_i}|y, z_{phy})}{\mathcal{P}(z_{n_i}|z_{phy})},
\end{equation}
where $n_i$ denotes the $i$-th type domain-specific noise (e.g., illumination, skin color, camera, motion, etc). Notably, the domain-specific noises are discrepant among different datasets. For example, head movement ($hm$) is rare in other datasets (e.g., UBFC-rPPG~\cite{UBFC}, PURE~\cite{PURE}, but abundant in the VIPL-HR dataset~\cite{rhythmnet}, which leads to the domain gap $\mathcal{P}_{train}(z_{hm}|y, z_{phy})\ll\mathcal{P}_{test}(z_{hm}|y, z_{phy})$, causing the poor generalization performance.

\begin{figure*}[!ht]
\begin{center}
\includegraphics[scale=0.85]{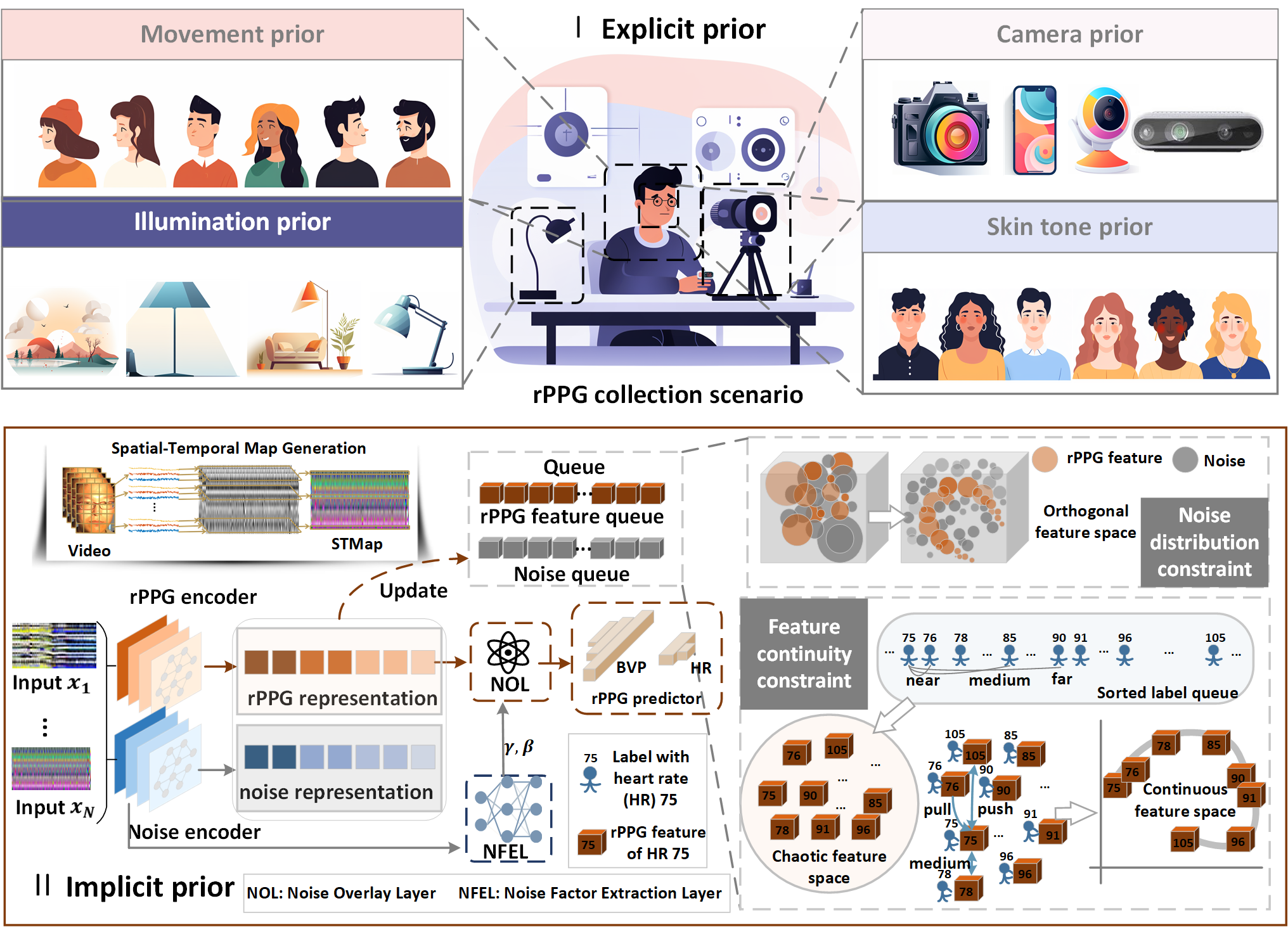}
\end{center}
\vspace{-2mm}
\caption{An overview of the proposed method. The above part shows the source and composition of the explicit prior in the collection process of the rPPG datasets. The following part shows the architecture of the entire two-flow network and how to constrain the rPPG feature and the implicit noise distribution, and finally inject the noise into the rPPG feature.}
\label{fig:Framework}
\vspace{-3mm}
\end{figure*}

\begin{figure*}[!ht]
 \begin{center}
\includegraphics[scale=1.3]{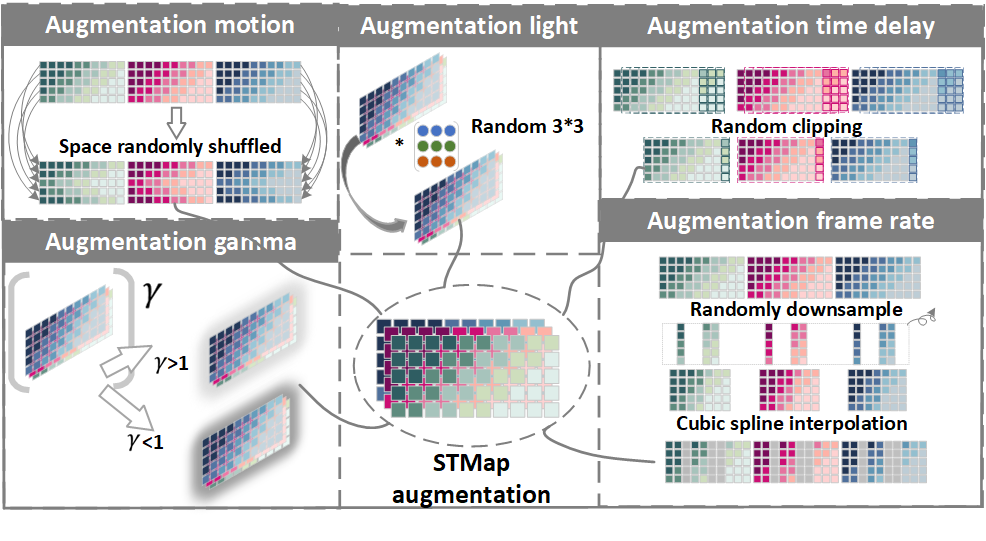}
\vspace{-2mm}
\captionof{figure}{Visualization of explicit priors. We visualized the five explicit augments mentioned in the Section \ref{sec:Explicit}, with different colors representing the three color channels: red, green, and blue. All the augmentation strategies are implemented on STMap.}
\label{fig:aug}
\end{center}%
\end{figure*}

\vspace{-3mm}
\subsection{Overall Framework}
\label{sec:overall_framwork}

To narrow the domain gap ($\mathcal{P}_{train}(z_{hm}|y, z_{phy})\ll\mathcal{P}_{test}(z_{hm}|y, z_{phy})$), we propose Greip framework to utilize both explicit prior and implicit prior, as shown in Figure \ref{fig:Framework}. Specifically, the input of our model is the spatial-temporal representation (STMap) extracted from face videos~\cite{rhythmnet, dual-GAN, NEST}. Then, multiple explicit priors are uniformly integrated into the network through STMap augmentation, which will be elaborated in the Section \ref{sec:Explicit}. The augmented STMap and original STMap are jointly defined as $\hat{\mathit{\text{ST}}}\in \mathbb{R}^{N\times T\times C}$, where $\it{N}$ denotes the number of ROIs, $\it{T}$ denotes the frames number of a video clip, $\it{C}$ denotes the number of channels($C=3$, including R, G and B), which is fed into a two-branch structure: 
\begin{equation} 
\begin{aligned}
    &z_{phy}^i,  z_{n}^i =\mathbf{E}_{rPPG}(\hat{\mathit{\text{ST}}}), \mathbf{E}_{noise}(\hat{\mathit{\text{ST}}}),\\
\end{aligned}
\end{equation}
where the encoders $\mathbf{E}_{rPPG}$ and $\mathbf{E}_{noise}$ are proposed to disentanle physiological and noise features, and $z_{phy}^i$ and $z_{n}^i$ denote the physiological and noise features, respectively. The implicit prior is used to disentangle and eliminate domain-specific noise $z_{n}$ in the latent feature distribution, which will be elaborated in the Section \ref{sec:Implicit}. Finally, the features are sent to the Noise Factor Extraction Layer (NFEL) and Noise Overlay Layer (NOL) for final heart rate and BVP signals.




\vspace{-3mm}
\subsection{Explicit Prior}
\label{sec:Explicit}

\begin{algorithm}[t]
\caption{Paradigm of explicit augmentation}
{\bfseries Input:} The original STMap $\mathit{\text{ST}}\in \mathbb{R}^{N\times T\times C}$; Random number $P$; Different augmentation probabilities: $P_{\gamma}$, $P_{f}$, $P_{t}$, $P_{l}$, $P_{m}$; $P_{\gamma} + P_{f} + P_{t}+ P_{l}+ P_{m} = 100\% $. \\
    { \small{1}\,\,\,:} $\textbf{for}$ iteration {\bfseries in} Max\_iterations: \\
    { \small{2}\,\,\,:} \ \ \ \, {\bfseries If} $P_{\gamma}>0 \& 0<P<P_{\gamma}$, add $\gamma$ augmentation:  \\
    { \small{3}\,\,\,:} \ \ \ \ \ \ \ \,$\text{ST}_{\gamma} = \mathcal{AUG}_{\gamma} (\text{ST})$ \\
    { \small{4}\,\,\,:} \ \ \ \,{\bfseries elif} $P_{f}>0 \& P_{\gamma}<P<1-P_{t}-P_{l}-P_{m}$, add frame rate augmentation:  \\
    { \small{5}\,\,\,:} \ \ \ \ \ \ \ \,$\text{ST}_{f} = \mathcal{AUG}_{f} (\text{ST})$  \\
    { \small{6}\,\,\,:} \ \ \ \,{\bfseries elif} $P_{t}>0 \& P_{\gamma} + P_{f}<P<1-P_{l}-P_{m}$, add time delay augmentation: \\
    { \small{7}\,\,\,:} \ \ \ \ \ \ \ \,$\text{ST}_{t} = \mathcal{AUG}_{t} (\text{ST})$  \\
    { \small{8}\,\,\,:} \ \ \ \,{\bfseries elif} $P_{l}>0 \& P_{\gamma} + P_{f}+P_{t}<P<1-P_{m}$, add light augmentation: \\
    { \small{9}\,\,\,:} \ \ \ \ \ \ \ \,$\text{ST}_{l} = \mathcal{AUG}_{l} (\text{ST})$  \\
    { \small{10}\,\,\,:}\ \ \ \,{\bfseries elif} $P_{m}>0 \& P_{\gamma} + P_{f}+P_{t} + p_{m} <P<1$, add motion augmentation: \\
    { \small{11}\,\,\,:} \ \ \ \ \ \ \ \,$\text{ST}_{m} = \mathcal{AUG}_{m} (\text{ST})$  \\  
    { \small{12}\,\,\,:} {\bfseries end for} \\
{\bfseries Output} The augmented STMap $\mathit{\text{ST}_{aug}}\in \mathbb{R}^{N\times T\times C} \in \{\text{ST}_{\gamma}, \text{ST}_{f}, \text{ST}_{t}, \text{ST}_{l}, \text{ST}_{m}\}$\\
\label{algorithm:explicit}
\end{algorithm}

The causes of noise in different data sets are systematically studied in this paper including 1) the camera, 2) the light source, 3) the skin tone, and 4) the head movement. To reduce the negative impact of likelihood $\mathcal{P}(z_{n}|y, z_{phy})$, we uniformly explicit Prior into the augmentation for STMap:

\textbf{Camera Prior}. When people use cathode ray tube CRT, they find that it has a problem: the regulation voltage is n times the original, and the corresponding screen luminance is not increased by n times, but a relationship similar to a power law curve. In order to make the brightness of the image displayed by the display equal to the brightness of the original object, it is necessary to gamma correct the brightness of the captured original image. The image data we can get is the computer stored data store, however, this store is gamma corrected data is not the original object data, so we need to undo the gamma correction. In fact, when we process images, we are doing it in linear space, and adding a nonlinear exponential simulation helps to simulate this noise. Specific implementation is as follows:
\begin{equation}
    \text{ST}_{\gamma}=(\text{ST})^{\gamma}, \gamma\in [0.8, 2.2],
\end{equation}
where ST is defined in the Section~\ref{sec:overall_framwork}. Inspired by \cite{chen2023deep_gamma}, we set $\gamma$ to a random number in [0.8, 2.2].

In addition, the frame rates of different cameras may also vary. For example, in the VIPL-HR dataset~\cite{rhythmnet}, the Logitech C310 camera has a frame rate of 25 fps, while the RealSense F200 camera has a frame rate of 30 fps. Even for the same camera, its frame rate may not be stable. Although the VIPL-HR dataset theoretically contains only two frame rates, 25 fps and 30 fps, we often observe other frame rates, such as 21 fps and 19 fps. It is common practice to sample them to uniform values, which will introduce noise. We will simulate this process by sampling:
\begin{equation}
    \text{ST}_f=\text{Cubic}(\text{Down}(S^{i,j})), i=0,1...C, j=0,1...N,
\end{equation}
where $s^{i,j}$ are one-dimensional pixel values from the STMap time domain.

Since the BVP sensor captures the physiological signal on the finger, and the camera captures the physiological signal on the human face, there is a certain time delay in the human blood flow from the finger to the face, so the rPPG signal of the face video is inherently different from the ground-truth. So, expanding the time domain to a certain extent is needed so that the video within the limit corresponds to the same ground-truth:
\begin{equation}
    \text{ST}_t=\text{Random}(S^{i,j}).
\end{equation}

\textbf{Light and skin color prior.} Owing to variations in the geographical origins of the collected datasets, a noticeable color bias exists among them. For instance, the UBFC-rPPG dataset~\cite{UBFC} comprises predominantly white subjects, whereas the VIPL-HR dataset~\cite{rhythmnet} consists entirely of individuals with a yellow complexion. In addition, the ambient light during the collection of different datasets will be different. These noises caused by skin color and illumination are represented in STMap as the differences in the overall pixel values of different chroma channels. In order to simulate this noise difference, we will dynamically weight RGB channels:
\begin{equation}
    \text{ST}_l=\left[
\begin{array}{c}
R_l \\
G_l \\
B_l
\end{array}
\right]
=
\left[
\begin{array}{ccc}
a_{11} & a_{12} & a_{13} \\
a_{21} & a_{22} & a_{23} \\
a_{31} & a_{32} & a_{33}
\end{array}
\right]
\left[
\begin{array}{c}
R \\ 
G \\
B
\end{array}
\right],
\end{equation}
where $a_{11}$... are random numbers between -0.5 and 0.5.

\textbf{Head motion prior.} The direct impact of head movement is that the detection key points are inaccurate, resulting in an inaccurate ROI division. We randomly interrupt the first dimension of STMap (representing the ROI region) to simulate this movement noise:
\begin{equation}
    \text{ST}_m=\text{Shuffle}(R_1, R_2, R_3..., R_N),
\end{equation}
where $R_i$ represents ROI region, the first dimension of STMap. 

In summary, we simulate five noises through explicit priors. These simulation strategies are applied in proportion to the source domain datasets so that the final noise distribution is similar to that of the target domain. To better understand explicit augmentation, we've added a visualization, as shown in Fig. \ref{fig:aug}, each part of the figure corresponds to a different explicit prior described above. This ratio depends on the specific noise differences between the selected target domain and the source domain, which is discussed in Sec~\ref{sec:further_study}. We conclude the whole explicit augmentation process in Aigorithm \ref{algorithm:explicit}.

\subsection{Implicit Prior}
\label{sec:Implicit}

\begin{algorithm}[t]
    \caption{Paradigm of implicit constraint}
{\bfseries Input:} The augmented STMap and original STMap are jointly defined as $\hat{\mathit{\text{ST}}}\in \mathbb{R}^{N\times T\times C}$; The ground truth label $L_{phy}$.
{\bfseries Initialize:} Ramdom initialize the queue: $Q_r\in \mathbb{R}^{K\times dim}$, $Q_n \in \mathbb{R}^{K\times dim}$, initialize the queue $Q_l \in \mathbb{R}^{K\times dim}$ with 75.\\
    { \small{1}\,\,\,:}  {\bfseries for } $i$ {\bfseries in} Max\_iterations:\\
    { \small{2}\,\,\,:} \ \ \ \ \ \,$z_{phy}^i,  z_{n}^i =\mathbf{E}_{rPPG}(\hat{\mathit{\text{ST}}}), \mathbf{E}_{noise}(\hat{\mathit{\text{ST}}})$ \\
    { \small{3}\,\,\,:} \ \ \ \ \ \,$\mathcal{L}_{con} \gets z_{phy}^i, L_{phy}^i, Q_{r},  Q_{l}$ \\
    { \small{4}\,\,\,:} \ \ \ \ \ \,$\mathcal{L}_{ort} \gets z_{n}^i, Q_{r}$ \\
    { \small{5}\,\,\,:} \ \ \ \ \ \,{\bfseries update $Q_r$} with $z_{phy}^i$\\
    { \small{6}\,\,\,:} \ \ \ \ \ \,{\bfseries update $Q_n$} with $z_{n}^i$\\
    { \small{7}\,\,\,:} \ \ \ \ \ \,{\bfseries update $Q_l$} with $L_{phy}^i$\\
    { \small{8}\,\,\,:} {\bfseries end for}	
\label{algorithm:implicit}	
\end{algorithm}

To more effectively counteract the influences of unknown noises, we introduce an implicit prior component designed to simulate the noise distribution in depth. Our underlying intuition is that by learning a pure rPPG feature representation, we can constrain the noise distribution to be orthogonal to the rPPG feature space. This approach allows us to minimize the rPPG information content while maximizing the noise content within the noise distribution. We posit that a robust rPPG feature should be resistant to noise. Hence, we fuse the rPPG feature with noise in order to maintain accurate predictions of the BVP signal and heart rate value. To actualize this concept, we first need to 1) spatially constrain the rPPG features and then 2) construct an orthogonal space.

\textbf{rPPG Feature Continuity Constraint.} It's crucial to note that rPPG, unlike other classification tasks, the heart rate is one of the labels for the rPPG task. For instance, a dataset commonly comprises continuous labels such as 75, 76, 77,..., 85. To make the most of this continuity characteristic, we should constrain the rPPG features within the network, making them continuous in the feature space. This approach assists the network in extracting pure rPPG features. Nonetheless, there is a limitation: the network can only output a feature of batch size, which falls short for the entire rPPG task. Consequently, we need to maintain a queue to ensure that it contains an adequate amount of rPPG features. Then, we simply need to use the distance between labels to constrain the network's output of rPPG features, and continuously update the queue to enable the network to learn a continuous and compact rPPG feature distribution. The constraints we impose are as follows:
\begin{equation}
\begin{aligned}
    &\mathcal{L}_{con}=-\sum\limits_{i=1}^{N}\sum\limits_{j=1}^{K}\frac{exp(w_{i,j})/v}{\sum_{k=1}^{K}exp(w_{i,k})/v}log\frac{exp(s_{i,j})}{\sum_{k=1}^{K}exp(s_{i,k})},\\
    &w_{i,j}=-|L_i-L_j|, s_{i,j}=sim(z_{phy}^i, Q_r^j),
\end{aligned}
\end{equation}
where $w_{i,j}$ represents the weight calculated by labels, the closer the labels are the higher the weight. $sim(z_r^i, Q_r^j$ denotes the cosine similarity of rPPG representations between batch ($z_r$) and the queue ($Q_r$), $N$ is the size of one batch, $K$ is the length of the rPPG feature queue. $v$ is the temperature constant. Therefore, we can maintain the distance of the rPPG representation in the feature space through the distance of the label to create a continues feature space.

\textbf{Noise distribution constaint.} We add a constraint to the noise such that the noise feature space is orthogonal to the rPPG feature space. 
\begin{equation}
\label{eq: t}
\begin{aligned}
    \mathcal{L}_{ort}=\frac{1}{3}(\sum\limits_{i=1}^N\sum\limits_{j=1}^KMSE(sim(z_n^i, Q_r^j),0),\\
    MSE(z_n, 1) + MSE(Q_r, 1)),  \mathcal{L}_{ort} > t,
\end{aligned}
\end{equation}
where $MSE(\cdot)$ denotes Mean Square Error. To make the training more stable, we added the last two normalized noise features and the rPPG queue. Since rPPG features and noise are not completely orthogonal in space, we add a constant $t$ to constraint $\mathcal{L}_{ort}$ is not too small.

\textbf{Queue.} The update of the queue follows the first-in-first-out principle. Its size is [K, dim], where K is the number of samples contained in the queue, and dim represents each sample feature dimension. The queue update process and the calculation of implicit constraint loss function are shown in Algorithm \ref{algorithm:implicit}.

\textbf{Fusion.} To promote dynamic features extraction, we adopt the Adaptive Instance Normalization (AdaIN) method~\cite{huang2017arbitrary_AdaIN}.
\begin{equation}
    \text{AdaIN}(x, \gamma, \beta) = \gamma(\frac{x- \mu(x)}{\sigma(x)}) + \beta,
\end{equation}
where $x$ is content input, $\mu(\cdot)$ and $\sigma(\cdot)$ represent channel-wise mean and standard deviation respectively, $\gamma$ and $\beta$ are affine parameters generated from the style input $y$. Here, when we use AdaIN, $x$ is replaced by rPPG feature and $y$ is replaced by noise.

The $\gamma$ and $\beta$ are calculated with MLP layers (NFEL), which reflect the feature distribution of noise representation. Then, the $\gamma$ and $\beta$ will be fused into the rPPG representation through NOL module which consisting of AdaIN layers and convolution layers with a residual connection, as follows:
\begin{equation}
\begin{aligned}
     \gamma, \beta&=\text{NFEL}(z_{n}^i)=\text{MLP}[\text{GAP}(z_{n}^i)],\\
    \textbf{Z}&=\text{ReLU}[\text{AdaIN}(K_1\otimes z_{phy}, \gamma, \beta)],\\
    \text{NOL}(z_{phy}, z_{n})&=\text{AdaIN}(K_2\otimes \textbf{Z}, \gamma, \beta) + z_{phy},
\end{aligned}
\end{equation}
where $K_1$ and $K_2$ are $3\times3$ convolution kernels, $\otimes$ is the convolution operation, and $\textbf{Z}$ is the intermediate variable.

 The fused features will be passed through a BVP prediction head and a heart rate (HR) prediction head respectively to obtain the final prediction results.  

\begin{table*}[!t] 
\small
\centering
\caption{A summary of the six public-domain datasets. C = Color Camera, N = NIR Camera, P = Smart Phone Frontal Camera; L = Lab Environment, D = Dim Environment, B = Bright Environment, G = Garage, CD = City Driving; E = Expression, S = Stable, SM = Slight Movement, LM = Large Movement, T = Talking; A = Asian, W = White, LH = Hispanic/Latino, AA = African American, I = Indian, C = Caucasian.}
\begin{tabular}{lcccc} 
\toprule  
\textbf{Dataset} & \textbf{Camera}& \textbf{Illumination variation } &\textbf{Head movement} & \textbf{Skin tone}\\
  \midrule 
  \textbf{VIPL-HR}~\cite{rhythmnet} & C/N/P & L/D/B & S/LM/T & A\\
  \textbf{PURE}~\cite{PURE} & C & L & S/SM/T & W\\
  \textbf{UBFC-rPPG}~\cite{UBFC} & C & L & S/SM/T & W/A\\
  \textbf{V4V}~\cite{V4V} & C & L & E & LH/W/AA/A\\
  \textbf{BUAA-MIHR}~\cite{BUAA} & C & D & S & A\\
  \textbf{MR-NIRP}~\cite{nowara2020near_MR_NIRP} & C/N & G/CD & S/SM/LM & I/C/A\\
\bottomrule 
\end{tabular} 
\label{tab:dataset}
\vspace{-3mm}
\end{table*} 

\subsection{rPPG predictor}
The rPPG predictor consists of a BVP predictor head and a heart rate predictor head, which are used to predict the BVP signal and heart rate value respectively. The output of the BVP prediction head is  $\mathit{\text{BVP}_{pre}}\in \mathbb{R}^{N\times L}$, and the output of the heart rate prediction head is $\mathit{\text{HR}_{pre}}\in \mathbb{R}^{N}$. Finally, the two predictors are constrained by two loss functions, $\mathcal{L}_{bvp}$ and $\mathcal{L}_{hr}$, respectively. The two loss functions are:


\begin{equation} \mathcal{L}_{bvp} = 1 - \frac{1}{N} \sum_{n=1}^{N} \frac{\sum_{l=1}^{L} (\text{gt}_{n,l} - \bar{\text{gt}}_{n}) (\text{pre}_{n,l} - \bar{\text{pre}}_{n})}{\sigma_{\text{gt}_n} \cdot \sigma_{\text{pre}_n} }, \end{equation}
where \(N\) and \(L\) represent the batch size and the length of the BVP signal, respectively. For a given segment of the BVP signal, \(\text{gt}_{n,l}\) denotes an individual value within that segment, while \(\bar{\text{gt}}_{n}\) denotes the average value of that BVP signal segment. Similarly, \(\text{pre}_{n,l}\) and \(\bar{\text{pre}}_{n}\) correspond to an individual predicted value and the average of the predicted BVP signal segment, respectively. \(\sigma_{\text{gt}n}\) is the standard deviation of \(\text{gt}_{n,l}\) over \(L\), \(\sigma_{\text{pre}_n}\) is the standard deviation of \(\text{pre}_{n,l}\) over \(L\). The entire loss function aims to compute the negative Pearson correlation coefficient between the predicted BVP signal and the ground truth BVP signal. 

\begin{equation}
\mathcal{L}_{hr} = \frac{1}{N} \sum_{i=1}^{N} \lvert \text{HR}_{pre}^i - \text{HR}_{gt}^i \rvert,
\end{equation}
where \(N\) denotes the batch size, while \(\text{HR}_{pre}^i\) and \(\text{HR}_{gt}^i\) represent the predicted and ground truth heart rate values, respectively. The function aims to compute the L1 loss, which quantifies the absolute difference between the predicted and actual heart rate values.

The ultimate loss function employed in training the entire network is as follows:

\begin{equation}
\mathcal{L}_{overall} = k_1\mathcal{L}_{bvp} + \lambda (k_2\mathcal{L}_{hr} + k_3\mathcal{L}_{con} + k_4\mathcal{L}_{ort}),
\end{equation}
where \(k_1\) to \(k_4\) serve as four trade-off parameters. To ensure stable training, we introduce an adaptation factor \(\lambda = \frac{2}{1 + \exp(-10 \cdot r)}\), where \( r = \frac{\text{iter}_{\text{current}}}{\text{iter}_{\text{total}}}\) represents the proportion of completed iterations relative to the total number of iterations. This adaptation factor is meticulously engineered to progressively integrate additional loss functions, with the exception of the \(\mathcal{L}_{bvp}\), into the optimization process of the network as the iteration progresses. This approach is instrumental in preserving the stability of the network's training.

\section{Experiments}
\label{sec:experiment}
\begin{table*}[!ht]
\centering
\small
\setlength{\tabcolsep}{0.0001mm}
\caption{HR estimation results on MSDG protocol. $^{*}$ means that these methods use the STMap as the input of CNN; $^{+}$ means that these methods are based on baseline (Rhythmnet~\cite{rhythmnet} without GRU). The best results are shown in bold. }
\begin{tabular}{cccccccccccccccc}
\toprule 
 & \multicolumn{3}{c}{\textbf{UBFC-rPPG}}             & \multicolumn{3}{c}{\textbf{PURE}}             & \multicolumn{3}{c}{\textbf{BUAA-MIHR}}             & \multicolumn{3}{c}{\textbf{VIPL-HR}}              & \multicolumn{3}{c}{\textbf{V4V}}               \\
\cmidrule(lr){2-4} \cmidrule(lr){5-7} 
\cmidrule(lr){8-10} \cmidrule(lr){11-13}  \cmidrule(lr){14-16} 
\textbf{Method} & \textbf{MAE↓}  & \textbf{RMSE↓} & \textbf{\quad r↑ \quad}    & \textbf{MAE↓}  & \textbf{RMSE↓} & \textbf{\quad r↑ \quad}    & \textbf{MAE↓}  & \textbf{RMSE↓} & \textbf{\quad r↑ \quad}    & \textbf{MAE↓}  & \textbf{RMSE↓}  & \textbf{\quad r↑ \quad}    & \textbf{MAE↓}  & \textbf{RMSE↓}  & \textbf{\quad r↑ \quad}    \\
\midrule 
\textbf{GREEN~\cite{GREEN}} & 8.02  & 9.18  & 0.36 & 10.32   & 14.27  & 0.52  & 5.82  & 7.99   & 0.56  & 12.18  & 18.23  & 0.25  & 15.64 & 21.43  & 0.06  \\
\textbf{CHROM~\cite{CHROM}} & 7.23 & 8.92 & 0.51 & 9.79 & 12.76  & 0.37 & 6.09 & 8.29 & 0.51 & 11.44 & 16.97   & 0.28 & 14.92  & 19.22  & 0.08          \\
\textbf{POS~\cite{POS}} & 7.35  & 8.04 & 0.49 & 9.82 & 13.44 & 0.34 & 5.04 & 7.12 & 0.63 & 14.59  & 21.26      & 0.19  & 17.65  & 23.22 & 0.04          \\
\midrule
\textbf{DeepPhys~\cite{deepphys}} & 7.82 & 8.42 & 0.54 & 9.34 & 12.56  & 0.55 & 4.78  & 6.74  & 0.69  & 12.56 & 19.13  & 0.14   & 14.52  & 19.11  & 0.14   \\
\textbf{TS-CAN~\cite{TS-CAN}} & 7.63 & 8.25  & 0.55   & 9.12 & 12.38  & 0.57 & 4.84 & 6.89 & 0.68  & 12.34  & 18.94  & 0.16   & 14.77 & 19.96   & 0.12          \\
\textbf{Rhythmnet$^*$~\cite{rhythmnet}}                                           & 5.79          & 7.91          & 0.78          & 7.39          & 10.49         & 0.77          & 3.38          & 5.17          & 0.84          & 8.97          & 12.16          & 0.49          & 10.16         & 14.57          & 0.34          \\
\textbf{Dual-GAN$^*$~\cite{dual-GAN}}                                           & 5.55          & 7.62          & 0.79          & 7.24          & 10.27         & 0.78          & 3.41          & 5.23          & 0.84          & 8.88          & 11.69          & 0.50          & 10.04         & 14.44          & 0.35          \\
\textbf{BVPNet$^*$~\cite{bvpnet}}                                           & 5.43          & 7.71          & 0.80          & 7.23           & 10.25         & 0.78          & 3.69          & 5.48          & 0.81          & 8.45          & 11.64          & 0.51          & 10.01         & 14.35          & 0.36          \\
\textbf{AD$^{*+}$~\cite{AD}}                                                 & 5.92          & 8.08          & 0.76          & 7.42          & 10.61         & 0.73          & 3.49          & 5.49          & 0.82          & 8.41          & 11.71          & 0.53          & 10.47         & 14.64          & 0.32          \\
\textbf{GroupDRO$^{*+}$~\cite{groupDRO}}                                           & 5.73          & 7.97          & 0.78         & 7.69          & 10.83         & 0.78          & 3.41          & 5.21          & 0.83             & 8.35          & 11.67          & 0.54          & 9.94          & 14.29          & 0.36          \\

\textbf{Coral$^{*+}$~\cite{coral}}                                           & 5.89          & 8.04          & 0.76          & 7.59          & 10.87         & 0.72          & 3.64          & 5.74          & 0.80          & 8.68          & 11.91          & 0.53          & 10.32         & 14.42          & 0.32      \\

\textbf{VREx$^{*+}$~\cite{VREX}}                                               & 5.59          & 7.68          & 0.81      & 7.24          & 10.14         & 0.78         & 3.27          & 5.01          & 0.86             & 8.37          & 11.62          & 0.54          & 9.82          & 14.16          & 0.37          \\
\textbf{NCDG$^{*+}$~\cite{NCDG}}                                               & 5.31          & 7.56          & 0.82          & 7.32          & 10.35         & 0.77          & 3.12          & 5.16          & 0.85          & 8.47          & 11.81          & 0.52          & 10.14         & 14.46          & 0.34          \\
\textbf{NEST$^{*+}$~\cite{NEST}}                                            & 4.67 & 6.79 & 0.86 & 6.71 & 9.59 & 0.81 & 2.88          & 4.69          & 0.89          & 7.86 & 11.15 & 0.58 & 9.27 & 13.79 & 0.41 \\
\midrule
\textbf{Baseline$^*$}                                           & 5.53         & 7.89          & 0.84          & 6.57          & 9.86         & 0.89          & 2.14          & 3.03          & 0.96          & 8.40          & 11.33          & 0.53          & 9.13         & 11.10          & 0.45          \\
\textbf{Greip$^{*+}$ w Ex-prior}  & 4.48          & 6.56          & 0.88          & 5.46          & 8.58          & 0.88          & 1.93   &2.60   & 0.97 & 7.39          & 10.78          & 0.63          & 8.90          & 10.73          & 0.46          \\
\textbf{Greip$^{*+}$ w Im-prior} & 4.72          & 6.89          & 0.85          & 5.64          & 9.05          & 0.87          & 1.78          & 2.48          & 0.97          & 8.07          & 11.33          & 0.54          & 8.81          & 11.31          & 0.50          \\
\textbf{Greip$^{*+}$ (Ours)}   & \textbf{4.08}          & \textbf{6.17}          & \textbf{0.88}          & \textbf{4.57}          & \textbf{7.71}          & \textbf{0.90}          & \textbf{1.69}          & \textbf{2.21}          & \textbf{0.98}          & \textbf{7.10}          & \textbf{10.31}          & \textbf{0.65}          & \textbf{8.46}          & \textbf{10.56}          & \textbf{0.53}           \\

\bottomrule 
\end{tabular}

\label{tab:MSDG}
\end{table*}

\subsection{Datasets and Metrics}
\textbf{Dataset.} The equipment, environment, motion disturbance and race of subjects used in the collection process of different datasets will be different, which will affect the generalization performance of rPPG method. We summarized these factors in the six datasets involved in the experiment in the Table~\ref{tab:dataset}, and described them more specifically as follows:

\textbf{VIPL-HR~\cite{rhythmnet}} have nine scenarios, three RGB cameras, different illumination conditions, and different levels of movement. \textbf{PURE~\cite{PURE}} contains 60 RGB videos from 10 subjects with six different activities, specifically, sitting still, talking, and four rotating and moving head variations. \textbf{UBFC-rPPG~\cite{UBFC}} containing 42 face videos with sunlight and indoor illumination. \textbf{V4V~\cite{V4V}} is designed to collect data with the drastic changes of physiological indicators by simulating ten tasks such as a funny joke, 911 emergency call, and odor experience. \textbf{BUAA-MIHR~\cite{BUAA}} is proposed to evaluate the performance of the algorithm against various illumination. \textbf{MR-NIRP~\cite{nowara2020near_MR_NIRP}} is a driving dataset with both NIR and RGB videos of a passenger’s face, along with pulse oximeter readings. We used the NIR portion for our experiments. This dataset, recorded in two driving scenarios - inside a garage and in city driving, presents various head motion and lighting conditions.

\textbf{Metrics}. Following methods~\cite{deepphys, dual-GAN, yu2022physformer}, standard deviation (SD), mean absolute error (MAE), root mean square error (RMSE), and Pearson’s correlation
coefficient ($r$) are used to evaluate the HR estimation. For the assessment of HRV measurements, which include low frequency (LF), high frequency (HF), and LF/HF, we employ MAE, RMSE, and $r$.


\subsection{Implemention Details}

The proposed method is implemented using Pytorch. The encoders $\mathbf{E}_{rPPG}$ and $\mathbf{E}_{noise}$ use ResNet 18 to extract the features. We set the batch size to 256 and the number of iterations to 40,000. The trade-off parameter \(k_1-k_4\) are set to 1, 0.1, 0.001, and 0.01 based on the scale of losses. The training process utilizes the Adam optimizer with a learning rate of 0.001. We use a default queue size of 5120. In each dataset, the STMap is sampled with a time window of 256, and the step of overlapping is set to 5. The network takes both the original STMap ($\mathit{\text{ST}^i}\in \mathbb{R}^{64\times 256\times 3}$) and the explicitly augmented STMap ($\mathit{\text{ST}_{aug}^i}\in \mathbb{R}^{64\times 256\times 3}$) as inputs.

\subsection{Multi-Source Domain Generalization}
\subsubsection{HR Estimation}
\begin{table*}[]
\setlength{\tabcolsep}{0.8mm}
\centering
\caption{HRV and HR estimation results on the MSDG protocol. e best results are shown in bold.}
\small
\begin{tabular}{clcccccccccccc}
\toprule
 &   & \multicolumn{3}{c}{\textbf{LF-(u.n)}}   & \multicolumn{3}{c}{\textbf{HF-(u.n)}}   & \multicolumn{3}{c}{\textbf{LF/HF}} & \multicolumn{3}{c}{\textbf{HR-(bpm)}}     \\
\cmidrule(lr){3-5} \cmidrule(lr){6-8} 
\cmidrule(lr){9-11} \cmidrule(lr){12-14}        
{\textbf{Target}}&{\textbf{Method}}  & \textbf{MAE↓}    & \textbf{RMSE↓}   & \textbf{\quad r↑ \quad}      & \textbf{MAE↓}    & \textbf{RMSE↓}   & \textbf{\quad r↑ \quad}     & \textbf{MAE↓}    & \textbf{RMSE↓}   & \textbf{\quad r↑ \quad}     & \textbf{MAE↓}    & \textbf{RMSE↓}   & \textbf{\quad r↑ \quad}     \\
\midrule 
\textbf{UBFC-rPPG} & 
\textbf{GREEN~\cite{GREEN}}    & 0.2355 & 0.2841 & 0.0924 & 0.2355 & 0.2841 & 0.0924 & 0.6695  & 0.9512 & 0.0467 & 8.0184  & 9.1776  & 0.3634 \\
& \textbf{CHROM~\cite{CHROM}}    & 0.2221 & 0.2817 & 0.0698 & 0.2221 & 0.2817 & 0.0698 & 0.6708  & 1.0542 & 0.1054 & 7.2291  & 8.9224  & 0.5123 \\
& \textbf{POS~\cite{POS}}      & 0.2364 & 0.2861 & 0.1359 & 0.2364 & 0.2861 & 0.1359 & 0.6515  & 0.9535 & 0.1345 & 7.3539  & 8.0402  & 0.4923 \\
& \textbf{NEST~\cite{NEST}}  & 0.0597 & 0.0782 & 0.2017 & 0.0597 & 0.0782 & 0.2017 & 0.2138  & 0.2824 & 0.3179 & 4.7471  & 6.8876  & 0.8546 \\
& \textbf{Greip (Ours)}  & \textbf{0.0583} & \textbf{0.0762} & \textbf{0.2516} & \textbf{0.0583} & \textbf{0.0762} & \textbf{0.2516} & \textbf{0.2085}  & \textbf{0.2714} & \textbf{0.3850} & \textbf{4.0869}  & \textbf{6.5038}  & \textbf{0.8596} \\
 \midrule 
\textbf{PURE} &\textbf{GREEN~\cite{GREEN}}   & 0.2539 & 0.3002 & 0.0326 & 0.2539 & 0.3002 & 0.0326 & 0.6525  & 0.8932 & 0.0417 & 10.3247 & 14.2693 & 0.4952 \\
& \textbf{CHROM~\cite{CHROM}}   & 0.2096 & 0.2751 & 0.1059 & 0.2096 & 0.2751 & 0.0759 & 0.5404  & 0.8266 & 0.1173 & 9.7914  & 12.7568 & 0.3732 \\
& \textbf{POS~\cite{POS}}      & 0.1959 & 0.2571 & 0.1684 & 0.1959 & 0.2571 & 0.1684 & 0.5373  & 0.846  & 0.1433 & 9.8273  & 13.4414 & 0.3432 \\
& \textbf{NEST~\cite{NEST}}  & 0.0635 & 0.0874 & 0.6422 & 0.0635 & 0.0874 & 0.6422 & 0.2255  & 0.3505 & 0.5734 & 7.6889  & 10.4783 & 0.7255 \\
& \textbf{Greip (Ours)}  & \textbf{0.0615} & \textbf{0.0835} & \textbf{0.6923} & \textbf{0.0615} & \textbf{0.0835} & \textbf{0.6923} & \textbf{0.2215}  & \textbf{0.3318} & \textbf{0.6714} & \textbf{6.8835}  & \textbf{10.1055} & \textbf{0.8364} \\
  \midrule                     
\textbf{BUAA-MIHR} & \textbf{GREEN~\cite{GREEN}}    & 0.3472 & 0.3951 & 0.0871 & 0.3472 & 0.3951 & 0.0871 & 0.6453  & 0.8632 & 0.0921 & 5.8231  & 7.9882  & 0.5624 \\
                      & \textbf{CHROM~\cite{CHROM}}    & 0.3786 & 0.3237 & 0.0682 & 0.3786 & 0.3237 & 0.0682 & 0.6813  & 0.8836 & 0.0715 & 6.0934  & 8.2938  & 0.5165 \\
                      & \textbf{POS~\cite{POS}}      & 0.3198 & 0.3762 & 0.0962 & 0.3198 & 0.3762 & 0.0962 & 0.6275  & 0.8424 & 0.1127 & 5.0407  & 7.1198  & 0.6374 \\
                      & \textbf{NEST~\cite{NEST}}  & 0.1436 & 0.1665 & 0.2955 & 0.1436 & 0.1665 & 0.2955 & 0.5514  & 0.6884 & 0.3004 & 3.3723  &5.8806  & 0.7647 \\
                      & \textbf{Greip (Ours)}  & \textbf{0.1285} & \textbf{0.1514} & \textbf{0.3665} & \textbf{0.1285} & \textbf{0.1514} & \textbf{0.3665} & \textbf{0.5029}  & \textbf{0.6334} & \textbf{0.4398} & \textbf{2.1500}  & \textbf{3.1842} & \textbf{0.9483} \\
\bottomrule 
\end{tabular}
\vspace{-3mm}
\label{HRV_DG}
\end{table*}
Following the experimental setup described in \cite{NEST}, we conducted our experiments on five datasets: UBFC-rPPG~\cite{UBFC}, PURE~\cite{PURE}, VIPL-HR~\cite{rhythmnet}, BUAA-MIHR~\cite{BUAA}, and V4V~\cite{V4V}. The current dataset was chosen as the target domain, while the remaining four datasets served as the source domain. We trained our model on the source domains and evaluated its performance on the target domain.To provide a comprehensive comparison, we compared our proposed method with three traditional algorithms, five deep learning methods, and five domain generalization methods. The detailed results and implementation strategies of these methods can be found in NEST~\cite{NEST} and will not be reiterated here.From the results presented in Table~\ref{tab:MSDG}, it is evident that our proposed method achieved significant improvements in the prediction results across all five datasets. Specifically, we observed an average improvement of approximately 1 bpm in terms of mean absolute error (mae) compared to the NEST method~\cite{NEST}. Furthermore, we analyzed the individual contributions of the explicit and implicit priors in our proposed method. It can be observed that when either part is added alone, there is an improvement in the results. Specifically, explicit priors had a significant enhancement effect on the VIPL-HR dataset, which consists of diverse lighting environments and motion scenes. These unique characteristics were not as prevalent in the other domains. By designing specific augmentations based on explicit priors, we were able to improve the performance on the VIPL-HR dataset more effectively than with implicit priors alone.

\subsubsection{HRV Estimation}
We utilized the HRV (LF, HR, LF/HF) index to evaluate the quality of the predicted BVP signal by measuring the low frequency, high frequency, and low frequency high frequency signal ratio. Due to the lack of a reliable ground-truth BVP signal in VIPL-HR and V4V datasets, we conducted this evaluation on the remaining three datasets (UBFC-rPPG~\cite{UBFC}, PURE~\cite{PURE}, BUAA-MIHR~\cite{BUAA}). Following the HR evaluation protocol, we treated the current dataset as the target domain and the other two datasets as the source domain. We trained the model on the source domain and tested it on the target domain. Overall, our proposed Greip method showed significant improvement on all three datasets. Accurate prediction of LF/HF and RF (Hz) is crucial as they reflect an individual's physical and cardiac activity status. This can aid in early disease screening, such as arrhythmia detection. Our approach achieved good performance on these metrics, thereby enhancing the potential of rPPG for widespread use.

\vspace{-3mm}
\subsection{Single-Source Domain Generalization}

To achieve model training on a single dataset with the aim of generalizing to unseen domains, we employed the Single-Source Domain Generalization (SSDG) approach. Specifically, we utilized the UBFC-rPPG dataset as our exclusive training source domain and selected the PURE and BUAA-MIHR datasets as target domains to thoroughly assess the generalization performance of our Greip model. Comparative results clearly demonstrate that Greip is capable of attaining its anticipated high performance levels, even when trained solely on a single dataset, as shown in Table~\ref{HR_SSDG}. This robust generalization capability can be attributed to two pivotal factors: 
Firstly, the explicit prior component of the model simulates a variety of noise conditions and translates them into corresponding data augmentation strategies, ensuring that the model maintains strong generalization performance even in the absence of target domain-specific noise types within the source domain.
Secondly, the implicit prior component effectively mitigates disturbances from unknown domains, enhancing the model's universal applicability across the remote photoplethysmography (rPPG) field.

\begin{table}[]
\small
\setlength{\tabcolsep}{0.1mm}
\centering
\caption{HR estimation results on SSDG protocol by training on the UBFC-rPPG. The best results are shown in bold.}
\begin{tabular}{ccccccc}
\toprule 
   & \multicolumn{3}{c}{\textbf{PURE}}             & \multicolumn{3}{c}{\textbf{BUAA-MIHR}}             \\
\cmidrule(lr){2-4} \cmidrule(lr){5-7} 
\textbf{Method}   & \textbf{MAE↓}  & \textbf{RMSE↓} & \textbf{\quad r↑ \quad}    & \textbf{MAE↓}  & \textbf{RMSE↓} & \textbf{\quad r↑ \quad}    \\
 \midrule
\textbf{GREEN~\cite{GREEN}}    & 10.32         & 14.27         & 0.52          & 5.82          & 7.99          & 0.56          \\
\textbf{CHROM~\cite{CHROM}}    & 9.79          & 12.76         & 0.37          & 6.09          & 8.29          & 0.51          \\

\textbf{POS~\cite{POS}}      & 9.82          & 13.44         & 0.34          & 5.04          & 7.12          & 0.63          \\
\textbf{NEST}~\cite{NEST}    & 6.07 & 9.06 & 0.76 & 2.56 &\textbf{2.73} & 0.78 \\
\midrule
\textbf{Greip (Ours)}  & \textbf{5.70} & \textbf{8.23} & \textbf{0.88} & \textbf{2.12} & 2.84 & \textbf{0.96} \\
\bottomrule
\end{tabular}
\label{HR_SSDG}
\vspace{-4mm}
\end{table}

\vspace{-4.5mm}
\subsection{Intra-Dataset Testing on VIPL-HR}
Following the protocol in \cite{dual-GAN, yu2022physformer, bvpnet, rhythmnet}, we evaluated the performance of the proposed method on the VIPL-HR dataset ~\cite{rhythmnet} using five-fold cross-validation. We compare the proposed method with four traditional methods (SAMC ~\cite{tulyakov2016self_SAMC}, POS~\cite{POS}, CHROM~\cite{CHROM}, I3D~\cite{carreira2017quo_I3D}) six DL-based methods (DeepPhys~\cite{deepphys}, BVPNet~\cite{bvpnet}, RhythmNet~\cite{rhythmnet}, CVD~\cite{niu2020video_CVD}, Physformer~\cite{yu2022physformer}, Dual-GAN~\cite{dual-GAN}), and two methods (NEST~\cite{NEST}, DOHA~\cite{sun2023resolve}) proposed for domain generalization. The results are from the corresponding papers. As shown in Table \ref{tab:VIPL}, the proposed method outperform all the SOTA methods. Indeed, the VIPL-HR dataset is complex and can be considered as a "multi-domain" dataset to some extent. Given its diverse lighting environments and motion scenes, our proposed cross-domain augmentation approach proved to be highly beneficial. The results obtained further support the effectiveness of our approach in addressing the challenges posed by this unique dataset.

\begin{table}[!t] 
\small
\centering
\caption{HR estimation results by our method and several state-of-the-art methods on the VIPL-HR dataset. The best results are shown in bold.}
\begin{tabular}{lcccccc} 
\toprule  
\textbf{Method} & \textbf{SD↓}& \textbf{MAE↓}& \textbf{RMSE↓} & \textbf{\quad r↑ \quad} \\
  \midrule 
  {\textbf{SAMC}~\cite{tulyakov2016self_SAMC}}  & 18.0 & 15.9 & 21.0  & 0.11 \\
  \textbf{POS}~\cite{POS}  & 15.3 & 11.5 & 17.2   & 0.30 \\
  \textbf{CHROM}~\cite{CHROM}& 15.1 & 11.4 & 16.9  & 0.28 \\
  \textbf{I3D}~\cite{carreira2017quo_I3D}  & 15.9 & 12.0 & 15.9  & 0.07 \\
  \textbf{DeepPhy}~\cite{deepphys} & 13.6 & 11.0 & 13.8  & 0.11 \\
  \textbf{BVPNet}~\cite{bvpnet} & 7.75 & 5.34 & 7.85   & 0.70 \\
  \textbf{RhythmNet}~\cite{rhythmnet} & 8.11 & 5.30 & 8.14  & 0.76 \\
  \textbf{CVD}~\cite{niu2020video_CVD}  & 7.92 & 5.02 & 7.97  & 0.79 \\
  \textbf{Physformer}~\cite{yu2022physformer}  & 7.74 & 4.97& 7.79  & 0.78\\
  \textbf{Dual-GAN}~\cite{dual-GAN}  & 7.63 & 4.93& 7.68  & 0.81\\
   \textbf{NEST}~\cite{NEST}  & 7.49  & 4.76  & 7.51  & 0.84 \\
    \textbf{DOHA}~\cite{sun2023resolve}  & - & 4.87  & 7.64  & 0.83 \\
  \midrule 
  \textbf{Baseline} & 8.62 & 5.50 & 8.65 & 0.78\\
  \textbf{Greip (Ours)}  & \textbf{7.11} & \textbf{4.71} & \textbf{7.12}   & \textbf{0.86} \\
 
\bottomrule 
\end{tabular} 
\label{tab:VIPL}
\vspace{-4mm}
\end{table}

\subsection{RGB to NIR}
Due to the scarcity of near-infrared (NIR) datasets in the rPPG field, the availability of additional physiological information, such as heart rate and cardiovascular activity, in NIR videos makes it crucial to explore the generalization from RGB to NIR. By achieving this generalization, we can extend the usability of existing RGB datasets to a wider range of applications. In this section, we present a novel contribution as we achieve, for the first time, the generalization from RGB datasets to NIR datasets. Our proposed model was trained on five RGB datasets (VIPL-HR~\cite{rhythmnet}, PURE~\cite{PURE}, UBFC-rPPG~\cite{UBFC}, V4V~\cite{V4V}, BUAA-MIHR~\cite{BUAA}) and tested on a NIR dataset (MR-NIRP~\cite{nowara2020near_MR_NIRP}).As depicted in Table \ref{tab:NIR}, our proposed method demonstrates a significant improvement over the Baseline, particularly with a reduction of nearly 6 bpm in mean absolute error (MAE). It is important to note that RGB videos capture information from the visible light spectrum, specifically the red, green, and blue wavelengths, while NIR videos capture information within the near-infrared spectrum. This fundamental difference in capturing and representing image information presents a significant challenge in generalizing from RGB datasets to NIR datasets. Our proposed method takes the first step in addressing this challenge, and the results highlight its effectiveness in this novel direction.

\begin{table}[!t] 
\small
\centering
\caption{HR estimation results by our method from RGB to NIR. The best results are shown in bold.}
\begin{tabular}{lcccccc} 
\toprule  
\textbf{Method} & \textbf{SD↓}& \textbf{MAE↓}& \textbf{RMSE↓} & \textbf{\quad r↑ \quad} \\
  \midrule 

   \textbf{Baseline}  & 11.81  & 21.00  & 23.77  &0.38 \\
  \textbf{Greip (Ours)}  & \textbf{11.16} & \textbf{14.12} & \textbf{16.96}   & \textbf{0.48} \\
\bottomrule 
\end{tabular} 
\label{tab:NIR}
\end{table}

\subsection{Self-supervised learning}
Following several self-supervised methods in rPPG filed~\cite{sun2022contrast,wang2022self,liu2023rppgMAE}, we conduct the self-supervised HR estimation experients on the proposed method on VIPL-HR dataset. In our self-supervised experiments, we removed the rPPG prediction head while keeping the remaining components unchanged. The methods we compared include four popular contrastive learning approaches ( MoCo~\cite{2020MoCo}, SIMSIAM~\cite{2021simsiam}, BOYL~\cite{grill2020byol}, SIMCLR~\cite{chen2020simclr}) and three self-supervised methods specifically designed for rPPG (Gideon21~\cite{gideon2021way}, Contrast-phys~\cite{sun2022contrast}, rPPG-MAE~\cite{liu2023rppgMAE}), as shown in Table~\ref{tab:linear-probe}. The results demonstrate that the proposed method performs exceptionally well even without the need for labels, surpassing the current state-of-the-art techniques.
\begin{table}[t]
    \centering
    \caption{Self-supervised HR estimation results of our
method and several state-of-the-art methods on VIPL-HR dataset. The best results are shown in bold.} 
    \label{tab:linear-probe}
    \resizebox{0.45\textwidth}{!} {\begin{tabular}{p{2.5cm} c c c} 
     \toprule
     \multirow{2}{*}{Method} &  \multicolumn{3}{c}{HR (bpm)}\\
     \cmidrule(lr){2-4}
     ~& \multicolumn{1}{c}{MAE $\downarrow$} & \multicolumn{1}{c}{RMSE $\downarrow$} & \multicolumn{1}{c}{$r$ $\uparrow$}  \\
     \midrule
     MoCo~\cite{2020MoCo}& 9.27 & 13.05 & 0.04  \\ 
     SIMSIAM~\cite{2021simsiam}& 8.43 & 11.73 & 0.14  \\ 
     BOYL~\cite{grill2020byol}& 8.98 & 12.43 & 0.08  \\ 
     SIMCLR~\cite{chen2020simclr}& 8.57 & 11.94 & 0.10  \\
     Gideon21~\cite{gideon2021way}& 9.80 & 15.48 & 0.38  \\
     Contrast-phys~\cite{sun2022contrast}& 8.55 & 12.65 & 0.40\\
    rPPG-MAE~\cite{liu2023rppgMAE}& 7.83 & 11.19 & 0.48  \\ 
    \midrule
    \textbf{Greip (Ours)} & \textbf{7.35} & \textbf{9.70} & \textbf{0.55} \\
     \bottomrule
     \end{tabular}}
\vspace{-5mm}
\end{table}

\vspace{-5mm}
\subsection{Semi-supervised learning}
The semi-supervised experiments reflect the dependency of the proposed method on labels during the training process. We incrementally increased the proportion of labeled data within the training dataset, as indicated in the Table~\ref{tab:semi}. Overall, the proposed method significantly outperforms rPPG-MAE. On a more granular level, with only 10 $\%$ of the training data being labeled, incorporating the remaining 90$\% $ of unlabeled data improved the MAE from 9.23 to 8.55. This suggests that the enhancement in Greip's performance largely relies on the data itself rather than the labels. This characteristic is attributable to the integration of prior knowledge, both explicit and implicit, within the network by Greip. As the proportion of labeled data in the training set increases, there is a corresponding improvement in the experimental outcomes. This is because the labels serve to correct the predictive direction of the network.
\begin{table}[t]\large
    \centering
    \caption{Semi-supervised HR estimation results of our method and one state-of-the-art method on VIPL-HR dataset.} 
    \label{tab:semi}
    \resizebox{0.45\textwidth}{!} {
    \begin{tabular} {p{2cm} p{3cm} p{3cm} c c c }
     \toprule
      \multirow{2}{*}{Method}& \multicolumn{1}{c}{Train Data} &  \multicolumn{1}{c}{Train Data} & \multicolumn{3}{c}{HR (bpm)}\\
      \cmidrule(lr){4-6}
     ~&\multicolumn{1}{c}{w. Label} &  \multicolumn{1}{c}{w/o Label}& \multicolumn{1}{c}{MAE $\downarrow$}& \multicolumn{1}{c}{RMSE $\downarrow$} & \multicolumn{1}{c}{$r$ $\uparrow$}\\
     \midrule
     \multirow{6}{*}{rPPG-MAE~\cite{liu2023rppgMAE}}&\multicolumn{1}{c}{10$\%$} &\multicolumn{1}{c}{/} & 9.40& 13.20 & 0.05\\
    ~& \multicolumn{1}{c}{10$\%$}  &\multicolumn{1}{c}{90$\%$} & 9.01& 12.69 & 0.35\\
     ~&\multicolumn{1}{c}{20$\%$} &\multicolumn{1}{c}{/} & 9.67& 13.70 & 0.04\\
     ~&\multicolumn{1}{c}{20$\%$} &\multicolumn{1}{c}{80$\%$} & 8.53& 12.19 & 0.35\\
     ~&\multicolumn{1}{c}{50$\%$}  &\multicolumn{1}{c}{/} & 8.08& 11.37 & 0.49\\
     ~&\multicolumn{1}{c}{50$\%$}  &\multicolumn{1}{c}{50$\%$ }& 6.54& 9.90 & 0.63\\
     \midrule
      \multirow{6}{*}{Greip (Ours) } &\multicolumn{1}{c}{10$\%$} &\multicolumn{1}{c}{/} & 9.23 & 11.96 & 0.11\\
    ~& \multicolumn{1}{c}{10$\%$}  &\multicolumn{1}{c}{90$\%$} & 8.55& 10.87 & 0.40\\
     ~&\multicolumn{1}{c}{20$\%$} &\multicolumn{1}{c}{/} & 8.98 & 11.20 & 0.23\\
     ~&\multicolumn{1}{c}{20$\%$} &\multicolumn{1}{c}{80$\%$} & 8.34& 10.35 & 0.50\\
     ~&\multicolumn{1}{c}{50$\%$}  &\multicolumn{1}{c}{/} & 7.67 & 9.68 & 0.56\\
     ~&\multicolumn{1}{c}{50$\%$}  &\multicolumn{1}{c}{50$\%$ }& 6.10 & 8.47  & 0.70 \\
     \bottomrule
     \end{tabular}
     }
\vspace{-5mm}
\end{table}

\begin{table*}[t]\Large
    \centering
    \caption{Cross-dataset evaluation for 3D mask face PAD between 3DMAD AND HKBU-MARSV1+.} 
    \label{tab:PAD}
    \resizebox{0.65\textwidth}{!} {\begin{tabular}{p{3cm} p{5cm} c c c c c c} 
     \toprule
     \multirow{2}{*}{\textbf{PAD Method}} & \multirow{2}{*}{\textbf{rPPG Method}} &  \multicolumn{3}{c}{\textbf{MARSV1+$\to$3DMAD}} &  \multicolumn{3}{c}{\textbf{3DMAD$\to$MARSV1+}}\\
     \cmidrule(lr){3-5} \cmidrule(lr){6-8}
     ~& ~ & \multicolumn{1}{c}{\textbf{HTER\_test $\downarrow$}} & \multicolumn{1}{c}{\textbf{EER $\downarrow$}} & \multicolumn{1}{c}{\textbf{AUC $\uparrow$}} & \multicolumn{1}{c}{\textbf{HTER\_test $\downarrow$}} & \multicolumn{1}{c}{\textbf{EER $\downarrow$}} & \multicolumn{1}{c}{\textbf{AUC $\uparrow$}} \\
     \midrule
    \multirow{3}{*}{\textbf{LrPPG~\cite{LrPPGliu20163d}}} & \textbf{CHROM~\cite{CHROM}} & 12.47 & 12.47 & 93.97 & 11.23 & 10.90 & 94.88 \\
    ~ & \textbf{ND-DeeprPPG~\cite{liu2023robust}} & 7.24 & 7.76 & 95.76 & 2.81 & 3.42 & 99.12 \\
    ~ & \textbf{Greip (Ours)} & \textbf{6.50} &  \textbf{6.65} & \textbf{96.20} & \textbf{2.50} & \textbf{3.30} & \textbf{99.15} \\  
    \midrule
    \multirow{3}{*}{\textbf{PPGSec~\cite{PPGSecnowara2017ppgsecure}}} & \textbf{CHROM~\cite{CHROM}} & 14.75 & 15.12 & 90.96 & 13.73 & 15.23 & 92.88 \\
    ~ & \textbf{ND-DeeprPPG~\cite{liu2023robust}} & 10.81 & 11.47 &  94.73 & 7.08 & 8.08 & 97.18 \\
    ~ & \textbf{Greip (Ours)} & \textbf{9.58} &  \textbf{10.24} & \textbf{95.30} & \textbf{6.78} & \textbf{7.56} & \textbf{99.50} \\  
     \bottomrule
     \end{tabular}}
\vspace{-3mm}
\end{table*}

\vspace{-3mm}
\subsection{rPPG for 3D Mask Presentation Attack Detection}
To rigorously evaluate the generalization performance of our proposed method, Greip, we applied it to the task of detecting 3D mask presentation attacks using remote photoplethysmography (rPPG) technology. Following the ND-DeeprPPG protocol~\cite{liu2023robust}, we initially pre-trained Greip on the COHFACE dataset~\cite{cohfaceheusch2017reproducible}. Subsequently, we employed two state-of-the-art face anti-spoofing techniques based on rPPG, namely LrPPG~\cite{LrPPGliu20163d} and PPGSec~\cite{PPGSecnowara2017ppgsecure}, to leverage the rPPG signals extracted by Greip for distinguishing 3D mask attacks. Our cross-dataset experiments were conducted on the 3DMAD~\cite{3dmaderdogmus2014spoofing} and HKBU-MARsV1+~\cite{marsvliu2018remote} datasets.

As demonstrated by our experimental results in Table \ref{tab:PAD}, Greip outperformed the traditional CHROM~\cite{CHROM} method and the contemporary ND-DeeprPPG~\cite{liu2023robust} approach in the task of 3D mask attack detection. This achievement can be attributed to the robust adaptation of Greip to various types of noise present in rPPG datasets. Although 3DMAD and HKBU-MARsV1+ are not specifically designed for rPPG tasks, as facial datasets, the noise characteristics they exhibit are similar to those in rPPG datasets. This confirms that Greip can effectively transfer to other downstream tasks based on rPPG. Not only does this highlight the exceptional generalization capability of Greip, but it also opens new avenues for future research in rPPG-based face liveness detection.

\vspace{-3mm}
\subsection{Further Study}
\label{sec:further_study}

\textbf{Impact of the augmentation strategies on different dataset.} It should be noted that the proposed augmentation strategies for different explicit priors are integrated into the overall rPPG task.  However, there may be certain biases for individual rPPG datasets.  Specifically, the noise levels and proportions of different types of noise vary in each dataset.  For example, the VIPL-HR dataset has more pronounced motion artifacts, and our data augmentation specifically targeting motion noise significantly improves the performance on this dataset.  However, the proportion of these noise levels cannot be quantified manually and can only be estimated through experiments.

\begin{figure}[]
\begin{center}
\includegraphics[scale=0.35]{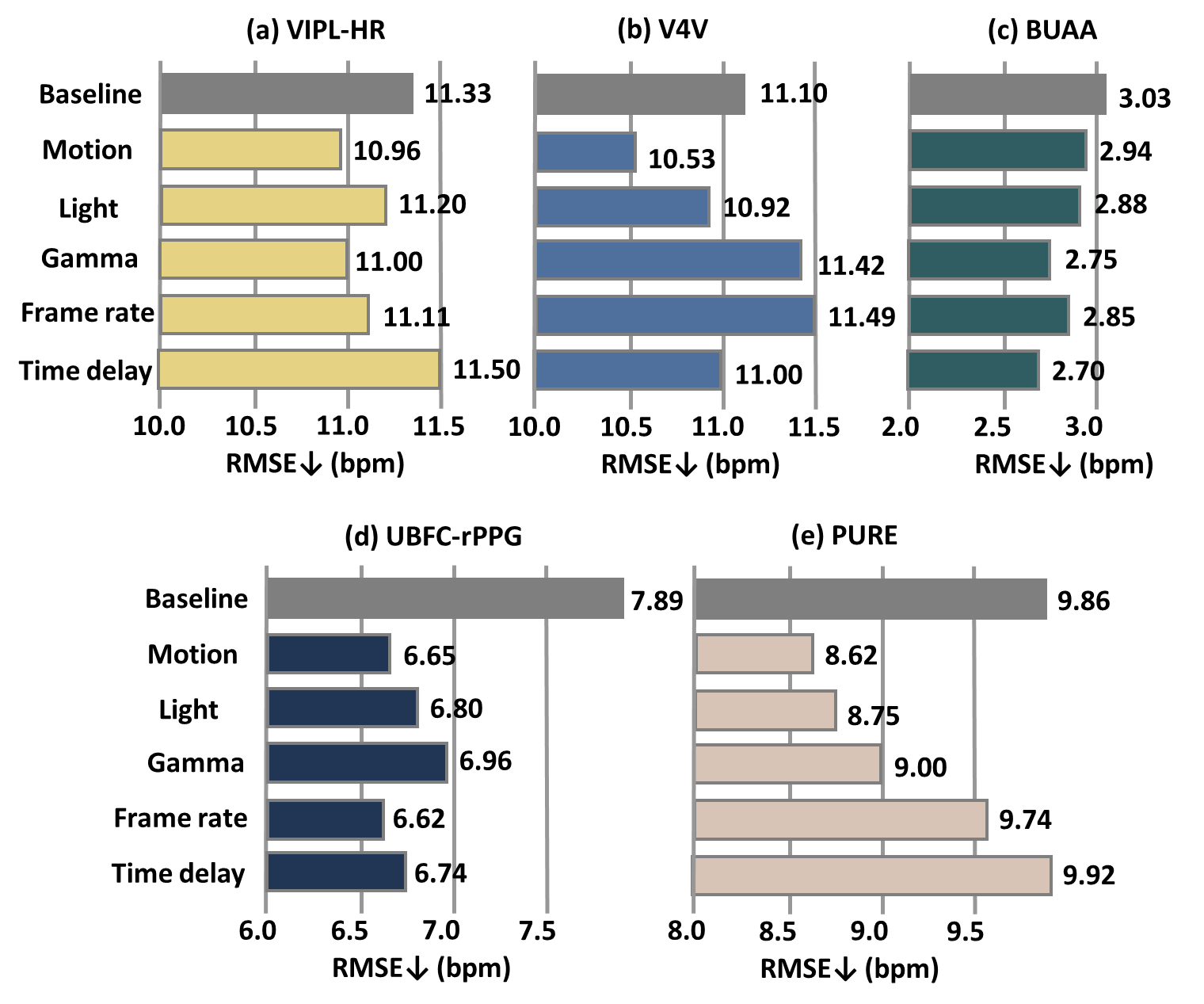}
\end{center}
\caption{The results of applying the proposed data augmentation on five datasets separately.}
\label{fig:aug_results}
\vspace{-5mm}
\end{figure}

In Figure \ref{fig:aug_results}, we separately apply one augmentation method to the target dataset and compare the effects of different augmentation methods.  It can be observed that different augmentation methods have varying effects on performance improvement individually, and there may even be cases where a particular augmentation method worsens the results.  Based on the analysis and results mentioned above, we apply different types and proportions of augmentation methods for different target datasets to ensure the effectiveness of the added data augmentation. We quantified this into the following formula:

\begin{equation}
\begin{aligned}
     ST_{aug} = &P_m\times ST_m + P_l\times ST_l + P_\gamma\times ST_{\gamma} + \\
    &P_f\times ST_f + P_t\times ST_t,
\end{aligned}
\end{equation}
where $P_*$ represents the proportion of different augmentation types, $ST_*$ denotes the augmentation method proposed for different explicit priors, as mentioned in Section \ref{sec:Explicit}.

According to the results in Figure \ref{fig:aug_results}, We add different proportions of augmentation to different target datasets. In this context, a ratio of 0\% represents that the content of this type of noise in the source domain is similar to that in the target domain, and continued addition will worsen the results. The settings for the other ratios are based on the corresponding results in Figure \ref{fig:aug_results}. When acting alone, the greater the improvement in generalization performance, the higher the ratio, and vice versa.

\begin{table}[!t] 
\small
\centering
\caption{The implementation probability of the proposed data augmentation on five datasets.}
\begin{tabular}{lcccccc} 
\toprule  
\textbf{Dataset} & \textbf{$P_m$} & \textbf{$P_l$} & \textbf{$P_\gamma$} & \textbf{$P_f$} & \textbf{$P_t$} \\
  \midrule 
  \textbf{VIPL-HR} & 30\% & 20\% & 30\%  & 20\% & 0\%\\
  \textbf{V4V}  & 40\% & 30\% & 0\%   & 0\% & 30\% \\
  \textbf{BUAA-MIHR}& 15\% & 15\% & 25\%  & 15\% & 30\% \\
  \textbf{UBFC-rPPG}  & 20\% & 15\% & 10\%  & 40\% & 15\% \\
  \textbf{PURE} & 40\% & 30\% & 20\%  & 10\% & 0\%\\
\bottomrule 
\end{tabular} 
\label{tab:add}
\end{table}

\begin{figure}[!t]
\begin{center}
\includegraphics[scale=0.9]{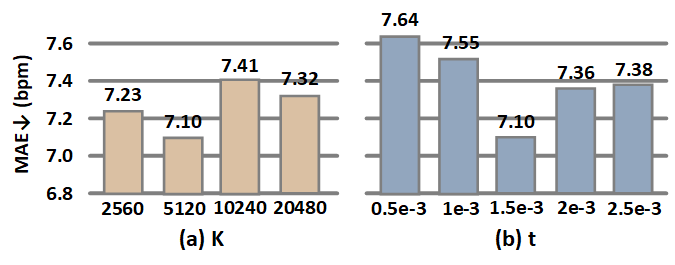}
\caption{Impacts of the hyperparameter (a) K and (b) t of the proposed method. }
\label{fig:ablation}
\end{center}
\vspace{-8mm}
\end{figure}


\textbf{Impact of $K$ in Greip.} The hyperparameter $K$ represents the size of the rPPG feature queue. The purpose is to maintain a queue that has a maximum number of rPPG features with different labels. This allows the model to be updated by calculating the feature distance from the rPPG features of the current batch. However, it is not necessarily true that a larger queue is always better. In fact, when the queue reaches an appropriate value (5120, as shown in figure \ref{fig:ablation} (a) ), it already contains enough rPPG features to effectively distinguish similarity. If the queue continues to increase, the performance will actually decrease. This is because a larger queue size causes the feature clusters belonging to a certain label to become too large, which blurs the boundaries between different label feature clusters and results a large overlap between feature clusters. Consequently, it becomes more difficult to distinguish these clusters in space.

\textbf{Impact of $t$ in Greip.} The hyperparameter $t$ is mentioned in Eq. \ref{eq: t}. The basic principle of orthogonal loss is based on the assumption that the noise space is completely orthogonal to the rPPG feature space, which is impossible to achieve in reality. Therefore, it is necessary to specify a value to limit the orthogonal loss so that it is not too small. From the Figure \ref{fig:ablation} (b), it can be observed that when $t$ is 0.5e-3, the results are poor. As $t$ increases, the results gradually improve until t reaches its lowest value at 1.5e-3. However, the results start to deteriorate again as t continues to increase. We suspect that setting a very small value for the orthogonal loss may cause the model to converge in a direction where the noise space is entirely orthogonal to the rPPG space, which is not realistic. This could lead to a reduction in the effectiveness of the continuous space constraints and ultimately harm the model's performance.
\begin{figure}[!t]
\begin{center}
\includegraphics[scale=0.85]{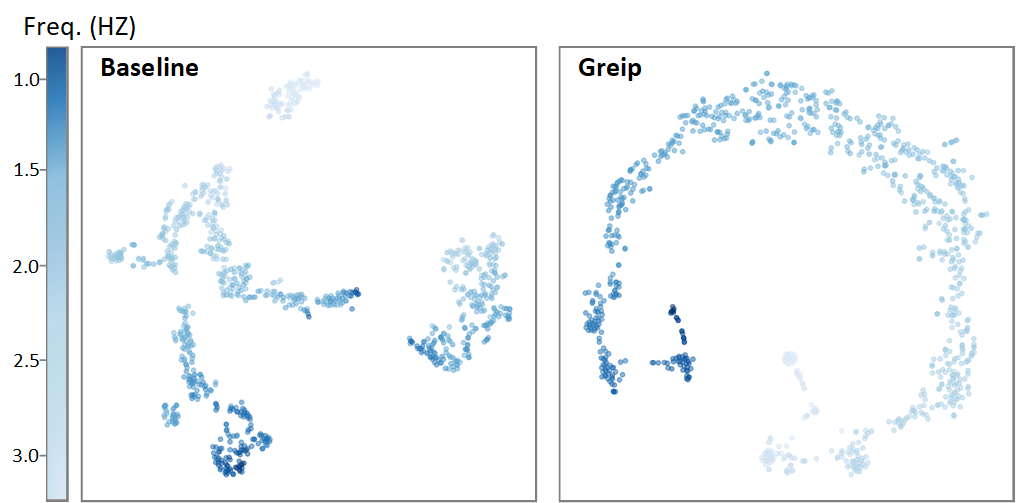}
\caption{Visualization of the rPPG feature. The heart rate value represented by the feature increases as the color lightens.} 
\label{fig:conti}
\end{center}
\vspace{-6mm}
\end{figure}

\textbf{Visualization of the rPPG feature.} We compared the rPPG characteristics of the baseline and Greip by visualizing them in Figure \ref{fig:conti}. The figure clearly demonstrates that the rPPG feature in Greip exhibits a smooth and continuous arc in space as the heart rate increases. In contrast, the rPPG feature obtained from the baseline training appears to be intermittent and irregular. This observation highlights that our spatial feature constraint aids the network in learning the continuity of heart rate prior to prediction. This continuity is represented by the smoothness of the rPPG feature, which in turn facilitates the extraction of relatively pure rPPG features and improves the accuracy of predictions.

\vspace{-0.8em}
\section{Conclusion}
\label{sec:conclusion}
In this paper, we propose a noval domain generalization method for rPPG task, named \textbf{Greip}, which integrate the explicit and implicit prior knowledge. Among them, explicit priors include camera prior, lighting prior, motion prior, and skin color prior. We design corresponding data augmentations to simulate these domain noises. We utilize the rPPG-specific label association constraint network to learn rPPG features and construct a continuous rPPG feature space. Additionally, we construct a noise space orthogonal to the rPPG feature space, combining the two to achieve implicit augmentation. Moreover, we also conduct a cross-modal domain generalization (RGB to NIR) for the first time. In the future, rPPG domain generalization will persist as a significant focal point of research. The emerging challenge is to devise methods that enhance cross-modal domain generalization, marking a new and exhilarating frontier in this field.

\ifCLASSOPTIONcaptionsoff
  \newpage
\fi

\bibliographystyle{IEEEtran}
\bibliography{IEEEabrv,reference}

\begin{thebibliography}{10}
\providecommand{\url}[1]{#1}
\csname url@samestyle\endcsname
\providecommand{\newblock}{\relax}
\providecommand{\bibinfo}[2]{#2}
\providecommand{\BIBentrySTDinterwordspacing}{\spaceskip=0pt\relax}
\providecommand{\BIBentryALTinterwordstretchfactor}{4}
\providecommand{\BIBentryALTinterwordspacing}{\spaceskip=\fontdimen2\font plus
\BIBentryALTinterwordstretchfactor\fontdimen3\font minus \fontdimen4\font\relax}
\providecommand{\BIBforeignlanguage}[2]{{%
\expandafter\ifx\csname l@#1\endcsname\relax
\typeout{** WARNING: IEEEtran.bst: No hyphenation pattern has been}%
\typeout{** loaded for the language `#1'. Using the pattern for}%
\typeout{** the default language instead.}%
\else
\language=\csname l@#1\endcsname
\fi
#2}}
\providecommand{\BIBdecl}{\relax}
\BIBdecl

\bibitem{GREEN}
W.~Verkruysse, L.~O. Svaasand, and J.~S. Nelson, ``Remote plethysmographic imaging using ambient light.'' \emph{Optics express}, vol.~16, no.~26, pp. 21\,434--21\,445, 2008.

\bibitem{poh2010non}
M.-Z. Poh, D.~J. McDuff, and R.~W. Picard, ``Non-contact, automated cardiac pulse measurements using video imaging and blind source separation.'' \emph{Optics express}, vol.~18, no.~10, pp. 10\,762--10\,774, 2010.

\bibitem{wang2017amplitude}
W.~Wang, A.~C. Den~Brinker, S.~Stuijk, and G.~De~Haan, ``Amplitude-selective filtering for remote-ppg,'' \emph{Biomedical optics express}, vol.~8, no.~3, pp. 1965--1980, 2017.

\bibitem{CHROM}
G.~De~Haan and V.~Jeanne, ``Robust pulse rate from chrominance-based rppg,'' \emph{IEEE Transactions on Biomedical Engineering}, vol.~60, no.~10, pp. 2878--2886, 2013.

\bibitem{wang2022self}
H.~Wang, E.~Ahn, and J.~Kim, ``Self-supervised representation learning framework for remote physiological measurement using spatiotemporal augmentation loss,'' in \emph{Proceedings of the AAAI Conference on Artificial Intelligence}, vol.~36, no.~2, 2022, pp. 2431--2439.

\bibitem{yang2022simper}
Y.~Yang, X.~Liu, J.~Wu, S.~Borac, D.~Katabi, M.-Z. Poh, and D.~McDuff, ``Simper: Simple self-supervised learning of periodic targets,'' \emph{arXiv preprint arXiv:2210.03115}, 2022.

\bibitem{liu2023rppgMAE}
X.~Liu, Y.~Zhang, Z.~Yu, H.~Lu, H.~Yue, and J.~Yang, ``rppg-mae: Self-supervised pre-training with masked autoencoders for remote physiological measurement,'' \emph{arXiv preprint arXiv:2306.02301}, 2023.

\bibitem{sun2024contrast}
Z.~Sun and X.~Li, ``Contrast-phys+: Unsupervised and weakly-supervised video-based remote physiological measurement via spatiotemporal contrast,'' \emph{IEEE Transactions on Pattern Analysis and Machine Intelligence}, 2024.

\bibitem{kurihara2021non}
K.~Kurihara, D.~Sugimura, and T.~Hamamoto, ``Non-contact heart rate estimation via adaptive rgb/nir signal fusion,'' \emph{IEEE Transactions on Image Processing}, vol.~30, pp. 6528--6543, 2021.

\bibitem{yang2021non_emotion_computing}
R.~Yang, Z.~Guan, Z.~Yu, X.~Feng, J.~Peng, and G.~Zhao, ``Non-contact pain recognition from video sequences with remote physiological measurements prediction,'' \emph{arXiv preprint arXiv:2105.08822}, 2021.

\bibitem{huang2021spatio_emotion_computing}
D.~Huang, X.~Feng, H.~Zhang, Z.~Yu, J.~Peng, G.~Zhao, and Z.~Xia, ``Spatio-temporal pain estimation network with measuring pseudo heart rate gain,'' \emph{IEEE Transactions on Multimedia}, vol.~24, pp. 3300--3313, 2021.

\bibitem{mcduff2014remote_emotion_computing}
D.~McDuff, S.~Gontarek, and R.~Picard, ``Remote measurement of cognitive stress via heart rate variability,'' in \emph{2014 36th annual international conference of the IEEE engineering in medicine and biology society}.\hskip 1em plus 0.5em minus 0.4em\relax IEEE, 2014, pp. 2957--2960.

\bibitem{speth2021deception}
J.~Speth, N.~Vance, A.~Czajka, K.~W. Bowyer, D.~Wright, and P.~Flynn, ``Deception detection and remote physiological monitoring: A dataset and baseline experimental results,'' in \emph{2021 IEEE International Joint Conference on Biometrics (IJCB)}.\hskip 1em plus 0.5em minus 0.4em\relax IEEE, 2021, pp. 1--8.

\bibitem{yu2021transrppg}
Z.~Yu, X.~Li, P.~Wang, and G.~Zhao, ``Transrppg: Remote photoplethysmography transformer for 3d mask face presentation attack detection,'' \emph{IEEE Signal Processing Letters}, vol.~28, pp. 1290--1294, 2021.

\bibitem{qi2020deeprhythm}
H.~Qi, Q.~Guo, F.~Juefei-Xu, X.~Xie, L.~Ma, W.~Feng, Y.~Liu, and J.~Zhao, ``Deeprhythm: Exposing deepfakes with attentional visual heartbeat rhythms,'' in \emph{Proceedings of the 28th ACM international conference on multimedia}, 2020, pp. 4318--4327.

\bibitem{Traditional_balakrishnan2013detecting}
G.~Balakrishnan, F.~Durand, and J.~Guttag, ``Detecting pulse from head motions in video,'' in \emph{Proceedings of the IEEE conference on computer vision and pattern recognition}, 2013, pp. 3430--3437.

\bibitem{Traditional_lam2015robust}
A.~Lam and Y.~Kuno, ``Robust heart rate measurement from video using select random patches,'' in \emph{Proceedings of the IEEE international conference on computer vision}, 2015, pp. 3640--3648.

\bibitem{Traditional_li2014remote}
X.~Li, J.~Chen, G.~Zhao, and M.~Pietikainen, ``Remote heart rate measurement from face videos under realistic situations,'' in \emph{Proceedings of the IEEE conference on computer vision and pattern recognition}, 2014, pp. 4264--4271.

\bibitem{Traditional_poh2010advancements}
M.-Z. Poh, D.~J. McDuff, and R.~W. Picard, ``Advancements in noncontact, multiparameter physiological measurements using a webcam,'' \emph{IEEE transactions on biomedical engineering}, vol.~58, no.~1, pp. 7--11, 2010.

\bibitem{2SR}
W.~Wang, S.~Stuijk, and G.~De~Haan, ``A novel algorithm for remote photoplethysmography: Spatial subspace rotation,'' \emph{IEEE transactions on biomedical engineering}, vol.~63, no.~9, pp. 1974--1984, 2015.

\bibitem{POS}
W.~Wang, A.~C. Den~Brinker, S.~Stuijk, and G.~De~Haan, ``Algorithmic principles of remote ppg,'' \emph{IEEE Transactions on Biomedical Engineering}, vol.~64, no.~7, pp. 1479--1491, 2016.

\bibitem{PVB}
G.~De~Haan and A.~Van~Leest, ``Improved motion robustness of remote-ppg by using the blood volume pulse signature,'' \emph{Physiological measurement}, vol.~35, no.~9, p. 1913, 2014.

\bibitem{bvpnet}
A.~Das, H.~Lu, H.~Han, A.~Dantcheva, S.~Shan, and X.~Chen, ``Bvpnet: Video-to-bvp signal prediction for remote heart rate estimation,'' in \emph{2021 16th IEEE International Conference on Automatic Face and Gesture Recognition (FG 2021)}.\hskip 1em plus 0.5em minus 0.4em\relax IEEE, 2021, pp. 01--08.

\bibitem{physnet}
Z.~Yu, X.~Li, and G.~Zhao, ``Remote photoplethysmograph signal measurement from facial videos using spatio-temporal networks,'' \emph{arXiv preprint arXiv:1905.02419}, 2019.

\bibitem{deepphys}
W.~Chen and D.~McDuff, ``Deepphys: Video-based physiological measurement using convolutional attention networks,'' in \emph{Proceedings of the european conference on computer vision (ECCV)}, 2018, pp. 349--365.

\bibitem{yu2022physformer}
Z.~Yu, Y.~Shen, J.~Shi, H.~Zhao, P.~H. Torr, and G.~Zhao, ``Physformer: Facial video-based physiological measurement with temporal difference transformer,'' in \emph{Proceedings of the IEEE/CVF conference on computer vision and pattern recognition}, 2022, pp. 4186--4196.

\bibitem{sun2022contrast}
Z.~Sun and X.~Li, ``Contrast-phys: Unsupervised video-based remote physiological measurement via spatiotemporal contrast,'' in \emph{European Conference on Computer Vision}.\hskip 1em plus 0.5em minus 0.4em\relax Springer, 2022, pp. 492--510.

\bibitem{dual-GAN}
H.~Lu, H.~Han, and S.~K. Zhou, ``Dual-gan: Joint bvp and noise modeling for remote physiological measurement,'' in \emph{Proceedings of the IEEE/CVF conference on computer vision and pattern recognition}, 2021, pp. 12\,404--12\,413.

\bibitem{TS-CAN}
X.~Liu, J.~Fromm, S.~Patel, and D.~McDuff, ``Multi-task temporal shift attention networks for on-device contactless vitals measurement,'' \emph{Advances in Neural Information Processing Systems}, vol.~33, pp. 19\,400--19\,411, 2020.

\bibitem{liu2023robust}
S.-Q. Liu and P.~C. Yuen, ``Robust remote photoplethysmography estimation with environmental noise disentanglement,'' \emph{IEEE Transactions on Image Processing}, 2023.

\bibitem{yu2023physformer++}
Z.~Yu, Y.~Shen, J.~Shi, H.~Zhao, Y.~Cui, J.~Zhang, P.~Torr, and G.~Zhao, ``Physformer++: Facial video-based physiological measurement with slowfast temporal difference transformer,'' \emph{International Journal of Computer Vision}, vol. 131, no.~6, pp. 1307--1330, 2023.

\bibitem{liu2023efficientphys}
X.~Liu, B.~Hill, Z.~Jiang, S.~Patel, and D.~McDuff, ``Efficientphys: Enabling simple, fast and accurate camera-based cardiac measurement,'' in \emph{Proceedings of the IEEE/CVF winter conference on applications of computer vision}, 2023, pp. 5008--5017.

\bibitem{chung2022domain1}
W.-H. Chung, C.-J. Hsieh, S.-H. Liu, and C.-T. Hsu, ``Domain generalized rppg network: Disentangled feature learning with domain permutation and domain augmentation,'' in \emph{Proceedings of the Asian Conference on Computer Vision}, 2022, pp. 807--823.

\bibitem{NEST}
H.~Lu, Z.~Yu, X.~Niu, and Y.-C. Chen, ``Neuron structure modeling for generalizable remote physiological measurement,'' in \emph{Proceedings of the IEEE/CVF Conference on Computer Vision and Pattern Recognition}, 2023, pp. 18\,589--18\,599.

\bibitem{sun2023resolve}
W.~Sun, X.~Zhang, H.~Lu, Y.~Chen, Y.~Ge, X.~Huang, J.~Yuan, and Y.~Chen, ``Resolve domain conflicts for generalizable remote physiological measurement,'' in \emph{Proceedings of the 31st ACM International Conference on Multimedia}, 2023, pp. 8214--8224.

\bibitem{du2023dual}
J.~Du, S.-Q. Liu, B.~Zhang, and P.~C. Yuen, ``Dual-bridging with adversarial noise generation for domain adaptive rppg estimation,'' in \emph{Proceedings of the IEEE/CVF Conference on Computer Vision and Pattern Recognition}, 2023, pp. 10\,355--10\,364.

\bibitem{rhythmnet}
X.~Niu, S.~Shan, H.~Han, and X.~Chen, ``Rhythmnet: End-to-end heart rate estimation from face via spatial-temporal representation,'' \emph{IEEE Transactions on Image Processing}, vol.~29, pp. 2409--2423, 2019.

\bibitem{HR-CNN}
R.~{\v{S}}petl{\'\i}k, V.~Franc, and J.~Matas, ``Visual heart rate estimation with convolutional neural network,'' in \emph{Proceedings of the british machine vision conference, Newcastle, UK}, 2018, pp. 3--6.

\bibitem{song2021pulsegan}
R.~Song, H.~Chen, J.~Cheng, C.~Li, Y.~Liu, and X.~Chen, ``Pulsegan: Learning to generate realistic pulse waveforms in remote photoplethysmography,'' \emph{IEEE Journal of Biomedical and Health Informatics}, vol.~25, no.~5, pp. 1373--1384, 2021.

\bibitem{gideon2021way}
J.~Gideon and S.~Stent, ``The way to my heart is through contrastive learning: Remote photoplethysmography from unlabelled video,'' in \emph{Proceedings of the IEEE/CVF international conference on computer vision}, 2021, pp. 3995--4004.

\bibitem{speth2023non}
J.~Speth, N.~Vance, P.~Flynn, and A.~Czajka, ``Non-contrastive unsupervised learning of physiological signals from video,'' in \emph{Proceedings of the IEEE/CVF Conference on Computer Vision and Pattern Recognition}, 2023, pp. 14\,464--14\,474.

\bibitem{data_man_shankar2018generalizing}
S.~Shankar, V.~Piratla, S.~Chakrabarti, S.~Chaudhuri, P.~Jyothi, and S.~Sarawagi, ``Generalizing across domains via cross-gradient training,'' \emph{arXiv preprint arXiv:1804.10745}, 2018.

\bibitem{data_man_yue2019domain}
X.~Yue, Y.~Zhang, S.~Zhao, A.~Sangiovanni-Vincentelli, K.~Keutzer, and B.~Gong, ``Domain randomization and pyramid consistency: Simulation-to-real generalization without accessing target domain data,'' in \emph{Proceedings of the IEEE/CVF International Conference on Computer Vision}, 2019, pp. 2100--2110.

\bibitem{AD}
Y.~Ganin and V.~Lempitsky, ``Unsupervised domain adaptation by backpropagation,'' in \emph{International conference on machine learning}.\hskip 1em plus 0.5em minus 0.4em\relax PMLR, 2015, pp. 1180--1189.

\bibitem{groupDRO}
G.~Parascandolo, A.~Neitz, A.~Orvieto, L.~Gresele, and B.~Sch{\"o}lkopf, ``Learning explanations that are hard to vary,'' \emph{arXiv preprint arXiv:2009.00329}, 2020.

\bibitem{wang2024inter}
M.~Wang, Y.~Liu, J.~Yuan, S.~Wang, Z.~Wang, and W.~Wang, ``Inter-class and inter-domain semantic augmentation for domain generalization,'' \emph{IEEE Transactions on Image Processing}, 2024.

\bibitem{meta_lv2022causality}
F.~Lv, J.~Liang, S.~Li, B.~Zang, C.~H. Liu, Z.~Wang, and D.~Liu, ``Causality inspired representation learning for domain generalization,'' in \emph{Proceedings of the IEEE/CVF Conference on Computer Vision and Pattern Recognition}, 2022, pp. 8046--8056.

\bibitem{sankaranarayanan2023meta}
S.~Sankaranarayanan and Y.~Balaji, ``Meta learning for domain generalization,'' in \emph{Meta-Learning with Medical Imaging and Health Informatics Applications}.\hskip 1em plus 0.5em minus 0.4em\relax Elsevier, 2023, pp. 75--86.

\bibitem{tang2022invariant}
K.~Tang, M.~Tao, J.~Qi, Z.~Liu, and H.~Zhang, ``Invariant feature learning for generalized long-tailed classification,'' in \emph{European Conference on Computer Vision}.\hskip 1em plus 0.5em minus 0.4em\relax Springer, 2022, pp. 709--726.

\bibitem{UBFC}
S.~Bobbia, R.~Macwan, Y.~Benezeth, A.~Mansouri, and J.~Dubois, ``Unsupervised skin tissue segmentation for remote photoplethysmography,'' \emph{Pattern Recognition Letters}, vol. 124, pp. 82--90, 2019.

\bibitem{PURE}
R.~Stricker, S.~M{\"u}ller, and H.-M. Gross, ``Non-contact video-based pulse rate measurement on a mobile service robot,'' in \emph{The 23rd IEEE International Symposium on Robot and Human Interactive Communication}.\hskip 1em plus 0.5em minus 0.4em\relax IEEE, 2014, pp. 1056--1062.

\bibitem{chen2023deep_gamma}
S.~Chen, S.~K. Ho, J.~W. Chin, K.~H. Luo, T.~T. Chan, R.~H. So, and K.~L. Wong, ``Deep learning-based image enhancement for robust remote photoplethysmography in various illumination scenarios,'' in \emph{Proceedings of the IEEE/CVF Conference on Computer Vision and Pattern Recognition}, 2023, pp. 6076--6084.

\bibitem{huang2017arbitrary_AdaIN}
X.~Huang and S.~Belongie, ``Arbitrary style transfer in real-time with adaptive instance normalization,'' in \emph{Proceedings of the IEEE international conference on computer vision}, 2017, pp. 1501--1510.

\bibitem{V4V}
A.~Revanur, Z.~Li, U.~A. Ciftci, L.~Yin, and L.~A. Jeni, ``The first vision for vitals (v4v) challenge for non-contact video-based physiological estimation,'' in \emph{Proceedings of the IEEE/CVF International Conference on Computer Vision}, 2021, pp. 2760--2767.

\bibitem{BUAA}
L.~Xi, W.~Chen, C.~Zhao, X.~Wu, and J.~Wang, ``Image enhancement for remote photoplethysmography in a low-light environment,'' in \emph{2020 15th IEEE International Conference on Automatic Face and Gesture Recognition (FG 2020)}.\hskip 1em plus 0.5em minus 0.4em\relax IEEE, 2020, pp. 1--7.

\bibitem{nowara2020near_MR_NIRP}
E.~M. Nowara, T.~K. Marks, H.~Mansour, and A.~Veeraraghavan, ``Near-infrared imaging photoplethysmography during driving,'' \emph{IEEE transactions on intelligent transportation systems}, vol.~23, no.~4, pp. 3589--3600, 2020.

\bibitem{coral}
B.~Sun and K.~Saenko, ``Deep coral: Correlation alignment for deep domain adaptation,'' in \emph{Computer Vision--ECCV 2016 Workshops: Amsterdam, The Netherlands, October 8-10 and 15-16, 2016, Proceedings, Part III 14}.\hskip 1em plus 0.5em minus 0.4em\relax Springer, 2016, pp. 443--450.

\bibitem{VREX}
D.~Krueger, E.~Caballero, J.-H. Jacobsen, A.~Zhang, J.~Binas, D.~Zhang, R.~Le~Priol, and A.~Courville, ``Out-of-distribution generalization via risk extrapolation (rex),'' in \emph{International Conference on Machine Learning}.\hskip 1em plus 0.5em minus 0.4em\relax PMLR, 2021, pp. 5815--5826.

\bibitem{NCDG}
C.~X. Tian, H.~Li, X.~Xie, Y.~Liu, and S.~Wang, ``Neuron coverage-guided domain generalization,'' \emph{IEEE Transactions on Pattern Analysis and Machine Intelligence}, vol.~45, no.~1, pp. 1302--1311, 2022.

\bibitem{tulyakov2016self_SAMC}
S.~Tulyakov, X.~Alameda-Pineda, E.~Ricci, L.~Yin, J.~F. Cohn, and N.~Sebe, ``Self-adaptive matrix completion for heart rate estimation from face videos under realistic conditions,'' in \emph{Proceedings of the IEEE conference on computer vision and pattern recognition}, 2016, pp. 2396--2404.

\bibitem{carreira2017quo_I3D}
J.~Carreira and A.~Zisserman, ``Quo vadis, action recognition? a new model and the kinetics dataset,'' in \emph{proceedings of the IEEE Conference on Computer Vision and Pattern Recognition}, 2017, pp. 6299--6308.

\bibitem{niu2020video_CVD}
X.~Niu, Z.~Yu, H.~Han, X.~Li, S.~Shan, and G.~Zhao, ``Video-based remote physiological measurement via cross-verified feature disentangling,'' in \emph{Computer Vision--ECCV 2020: 16th European Conference, Glasgow, UK, August 23--28, 2020, Proceedings, Part II 16}.\hskip 1em plus 0.5em minus 0.4em\relax Springer, 2020, pp. 295--310.

\bibitem{2020MoCo}
K.~He, H.~Fan, Y.~Wu, S.~Xie, and R.~Girshick, ``Momentum contrast for unsupervised visual representation learning,'' in \emph{Proceedings of the IEEE/CVF conference on computer vision and pattern recognition}, 2020, pp. 9729--9738.

\bibitem{2021simsiam}
X.~Chen and K.~He, ``Exploring simple siamese representation learning,'' in \emph{Proceedings of the IEEE/CVF conference on computer vision and pattern recognition}, 2021, pp. 15\,750--15\,758.

\bibitem{grill2020byol}
J.-B. Grill, F.~Strub, F.~Altch{\'e}, C.~Tallec, P.~Richemond, E.~Buchatskaya, C.~Doersch, B.~Avila~Pires, Z.~Guo, M.~Gheshlaghi~Azar \emph{et~al.}, ``Bootstrap your own latent-a new approach to self-supervised learning,'' \emph{Advances in neural information processing systems}, vol.~33, pp. 21\,271--21\,284, 2020.

\bibitem{chen2020simclr}
T.~Chen, S.~Kornblith, M.~Norouzi, and G.~Hinton, ``A simple framework for contrastive learning of visual representations,'' in \emph{International conference on machine learning}.\hskip 1em plus 0.5em minus 0.4em\relax PMLR, 2020, pp. 1597--1607.

\bibitem{LrPPGliu20163d}
S.~Liu, P.~C. Yuen, S.~Zhang, and G.~Zhao, ``3d mask face anti-spoofing with remote photoplethysmography,'' in \emph{Computer Vision--ECCV 2016: 14th European Conference, Amsterdam, The Netherlands, October 11--14, 2016, Proceedings, Part VII 14}.\hskip 1em plus 0.5em minus 0.4em\relax Springer, 2016, pp. 85--100.

\bibitem{PPGSecnowara2017ppgsecure}
E.~M. Nowara, A.~Sabharwal, and A.~Veeraraghavan, ``Ppgsecure: Biometric presentation attack detection using photopletysmograms,'' in \emph{2017 12th IEEE International Conference on Automatic Face \& Gesture Recognition (FG 2017)}.\hskip 1em plus 0.5em minus 0.4em\relax IEEE, 2017, pp. 56--62.

\bibitem{cohfaceheusch2017reproducible}
G.~Heusch, A.~Anjos, and S.~Marcel, ``A reproducible study on remote heart rate measurement,'' \emph{arXiv preprint arXiv:1709.00962}, 2017.

\bibitem{3dmaderdogmus2014spoofing}
N.~Erdogmus and S.~Marcel, ``Spoofing face recognition with 3d masks,'' \emph{IEEE transactions on information forensics and security}, vol.~9, no.~7, pp. 1084--1097, 2014.

\bibitem{marsvliu2018remote}
S.-Q. Liu, X.~Lan, and P.~C. Yuen, ``Remote photoplethysmography correspondence feature for 3d mask face presentation attack detection,'' in \emph{Proceedings of the European Conference on Computer Vision (ECCV)}, 2018, pp. 558--573.

\end{thebibliography}

\end{document}